\documentclass{article} 
\usepackage{iclr2025_conference,times}
\pdfoutput=1 

\usepackage{amsmath,amsfonts,bm}

\def\eqref#1{equation~\ref{#1}}

\def\1{\bm{1}}

\DeclareMathAlphabet{\mathsfit}{\encodingdefault}{\sfdefault}{m}{sl}
\SetMathAlphabet{\mathsfit}{bold}{\encodingdefault}{\sfdefault}{bx}{n}

\usepackage{hyperref}
\usepackage{url}

\usepackage{graphicx}
\usepackage{amsmath}
\usepackage{amssymb}
\usepackage{pifont}

\usepackage[nolist]{acronym}
\usepackage{tocloft}

\usepackage{dsfont}
\usepackage{tikz}
\usetikzlibrary{
graphs, calc, backgrounds, 
matrix, positioning, shapes.geometric,
decorations.pathreplacing,
fit
}
\usepackage{booktabs}
\usepackage{multirow}
\usepackage{bm}
\usepackage[inline]{enumitem}

\usepackage{pgfplots}
\pgfplotsset{compat=1.9} 
\usepackage{caption}
\usepackage{subcaption}
\usepackage{wrapfig}

\title{T-Graphormer: Using Transformers for Spatiotemporal Forecasting}

\iclrfinalcopy

\author{Hao Yuan Bai \\
School of Computer Science\\
McGill University\\
Mila, University of Montreal\\
Montreal, Quebec, Canada \\
\texttt{hao.bai@mail.mcgill.ca} \\
\And
Xue Liu \\
School of Computer Science\\
McGill University\\
Mila, University of Montreal\\
Montreal, Quebec, Canada \\
\texttt{xueliu@cs.mcgill.ca} \\
}

\begin{document}

\begin{acronym}[SimCLR]\itemsep0pt
    \acro{CNN}{Convolutional Neural Network}
    \acro{GAN}{Generative Adversarial Network}
    \acro{VAE}{Variational Autoencoder}
    \acro{ViT}{Vision Transformer}
    \acro{NLP}{Natural Language Processing}
    \acro{RNN}{Recurrent Neural Network}
    \acro{LSTM}{Long Short-Term Memory}
    \acro{SSM}{State-Space Model}
    \acro{GPT}{Generative Pre-trained Transformer}
    \acro{BERT}{Bidirectional Encoder Representations from Transformers}
    \acro{GloVe}{Global Vector}
    \acro{CBOW}{Continuous Bag-Of-Words}
    \acro{BEiT}{Bidirectional Encoder representation from Image Transformers}
    \acro{SimCLR}{Simple Framework for Contrastive Learning of Visual Representations}
    \acro{GNN}{Graph Neural Network}
    \acro{GCN}{Graph Convolutional Network}
    \acro{ARMA}{Autoregressive Moving-Average}
    \acro{ARIMA}{Autoregressive Integrated Moving-Average}
    \acro{SVR}{Support Vector Regression}
    \acro{DCRNN}{Diffusion Convolutional Recurrent Neural Network}
    \acro{ASTGCN}{Attention-based Spatial-Temporal Graph Convolutional Network}
    \acro{STGCN}{Spatio-Temporal Graph Convolutional Networks}
    \acro{STSGCN}{Spatial-Temporal Synchronous Graph Convolutional Network}
    \acro{MTGNN}{Multivariate Time Series Forecasting with Graph Neural Networks}
    \acro{STEP}{STGNN is Enhanced by scalable time series Pre-training model}
    \acro{GTS}{Graph for Time Series}
    \acro{AD}{Alzheimer’s Disease}
    \acro{MP}{Misfolded Proteins}
    \acro{ESM}{Epidemic Spreading Model}
    \acro{ADNI}{Alzheimer’s Disease Neuroimaging Initiative}
    \acro{A-beta}[A$\beta$]{Amyloid-$\beta$}
    \acro{MCI}{Mild Cognitive Impairment}
    \acro{APP}{Amyloid-$\beta$ Precursor Protein (APP)}
    \acro{T-Graphormer}{Temporal Graphormer}
    \acro{FFN}{Feed-Forward Network}
    \acro{GELU}[$\mathrm{GELU}$]{Gaussian Error Linear Unit}
    \acro{MAE}{Masked Autoencoders}
\end{acronym}

\maketitle

\begin{abstract}
Spatiotemporal data is ubiquitous, and forecasting it has important applications in many domains. However, its complex cross-component dependencies and non-linear temporal dynamics can be challenging for traditional techniques. Existing methods address this by learning the two dimensions separately. Here, we introduce \ac{T-Graphormer}, a Transformer-based approach capable of modelling spatiotemporal correlations simultaneously. By adding temporal encodings in the Graphormer architecture, each node attends to all other tokens within the graph sequence, enabling the model to learn rich spacetime patterns with minimal predefined inductive biases. We show the effectiveness of T-Graphormer on real-world traffic prediction benchmark datasets. Compared to state-of-the-art methods, T-Graphormer reduces root mean squared error (RMSE) and mean absolute percentage error (MAPE) by up to 20\% and 10\%.
\end{abstract}

\section{Introduction}
Time series data is prevalent across various disciplines and appears in different forms. In retail, it manifests as customer orders over time; in finance, as stock prices; in energy grid optimization, as electricity consumption or transformer temperature \citep{bose2017probabilistic}; and in geography, as geopotential or temperature measurements. Accurate prediction of time series has long been a critical problem with significant applications, leading to the development of many techniques such as spectral analysis, linear models, and state-space models \citep{brockwell2002introduction}. 

\begin{wrapfigure}{r}{0.5\textwidth}
    \centering
    \includegraphics[width=0.4\textwidth]{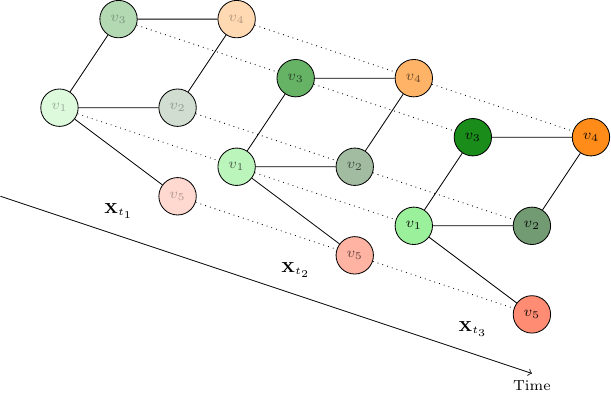}
    \caption{Visualization of a multivariate time series with graph structure. In this example, the graph remains static over time, similar to traffic networks. The node colours become more opaque to indicate more recent data points.}
    \label{fig:time_series}
\end{wrapfigure}

In this work, we focus on spatiotemporal data where each observation at time $t$ is a vector/matrix with components organized in a graph structure (Figure ~\ref{fig:time_series}). Traffic flow is a prime example of such data. While traffic networks might appear ``grid-like", their spatial structure is non-Euclidean. Two roads close in Euclidean space can exhibit different behaviours, yet roads far in Euclidean space could exhibit similar behaviours \citep{li2018diffusion}. For example, in a city like Los Angeles, two geographically close roads (a residential street and its parallel highway) experience vastly different traffic conditions due to differences in traffic volume, speed limits, and access points. Conversely, roads are far in Euclidean space (separate segments of the same highway) can exhibit similar behaviours if they are influenced by the same traffic flow dynamics. These complex spatial relationships combined with the non-stationarity of traffic flow have proven difficult for traditional machine learning techniques such as ARIMA and Kalman filtering \citep{okutani1984dynamic, lippi2013short, box2015time}.

Traffic prediction, similar to advancements seen in computer vision and \ac{NLP} \citep{NIPS2012_c399862d, he2016deep, radford2018improving, devlin2018bert}, has shifted from traditional machine learning methods to non-linear deep learning models. With improvements in data mining and computational resources, these models are increasingly capable of learning representations that capture complex dynamics. Many researchers have employed convolutional modules to address spatial and temporal dependencies separately \citep{li2018diffusion, yu2018spatio, wu2019graph, wu2020connecting}. By interleaving temporal and graph learning steps, these models exchange features learned in the time and space dimensions.


In this paper, we investigate whether directly modelling spatiotemporal data within a single unified framework leads to superior representation learning compared to approaches that use separate spatial and temporal learning steps. To do so, we explore variants of the Transformer architecture. Although originally designed for sequential data, Transformers can handle complex data structures through additional encoding techniques. For instance, architectures like Graphormer \citep{ying2021transformers} and \ac{ViT} \citep{dosovitskiy2020image, feichtenhofer2022masked} successfully adapt these techniques to recover structural information that is lost when the data is flattened and treated sequentially. 

Building on this foundation, we introduce T-Graphormer (Temporal-Graphormer). By extending the encoding techniques from Graphormer into the temporal dimension, T-Graphormer can leverage the global attention mechanism inherent in Transformers to capture spatial \textit{and} temporal relationships simultaneously. Our contributions are as follows:
\begin{enumerate}
    \item Our experiments on real-world traffic forecasting datasets demonstrate that T-Graphormer achieves state-of-the-art performance, surpassing existing methods by a significant margin.
    \item By analyzing its attention scores, we show that it learns spatial and temporal relationships without predefined structural biases.
    \item Through ablation studies, we identify which encoding methods in \ac{T-Graphormer} are responsible for its predictive abilities.
\end{enumerate}

\section{Related Works}\label{sec: related_works}
\subsection{Traffic Prediction}\label{sec: traffic_related}

As deep learning thrived in NLP and computer vision, many borrowed architectures from these fields for traffic forecasting. \citet{yu2018spatio} proposed \ac{STGCN}, which uses \ac{CNN} architectures to extract graphical and temporal features in traffic data. Each layer in \ac{STGCN} contains a ``sandwich" structure with two gated sequential convolution layers and a spatial graph convolution layer in between. Around the same time, \citet{li2018diffusion} developed \ac{DCRNN}. Inspired by the work by \citet{atwood2016diffusion}, it captures spatial dependencies using diffusion convolution, which models traffic flow as a diffusion process characterized by random walks. \ac{DCRNN} then captures temporal dependencies with \ac{RNN}. Subsequently, \citet{wu2019graph} created Graph WaveNet, which combines graph convolutional networks with the WaveNet architecture \citep{oord2016wavenet}. In WaveNet, dilated causal convolutions \citep{yu2016conv} are used to expand the receptive field for capturing historical temporal context. These convolutions are efficient, requiring fewer layers by skipping inputs based on a dilation factor that typically grows exponentially with each layer. Dilated convolutions are also utilized in \ac{MTGNN} by \citet{wu2020connecting}.

With the introduction of the attention mechanism in Transformers, many began integrating it into their spatial and temporal learning modules. For instance, \ac{ASTGCN} \citep{guo2019attention} integrates convolutional networks with temporal and spatial attention modules to selectively focus on critical information. Similarly, \citet{zheng2020gman} proposed GMAN, which uses an autoencoder architecture where each block integrates spatial and temporal attention mechanisms.

More recently, \citet{jiang2023pdformer} and \citet{liu2023spatio} introduced PDFormer and STAEformer, respectively. Both models utilize Transformer-based encoders for representation learning and fully connected layers for prediction. PDFormer incorporates domain knowledge into its attention heads, whereas STAEformer employs learnable spatiotemporal embeddings. Importantly, both models still rely on separate spatial and temporal learning modules. In PDFormer, the attention heads are split for different data dimensions (data is indexed by time and space), and in STAEformer, the input sequence is reshaped between time and space before entering the encoder.

\begin{figure}[tb]
    \centering
    \begin{subfigure}[b]{0.4\linewidth}
        \centering
        \includegraphics[width=\linewidth]{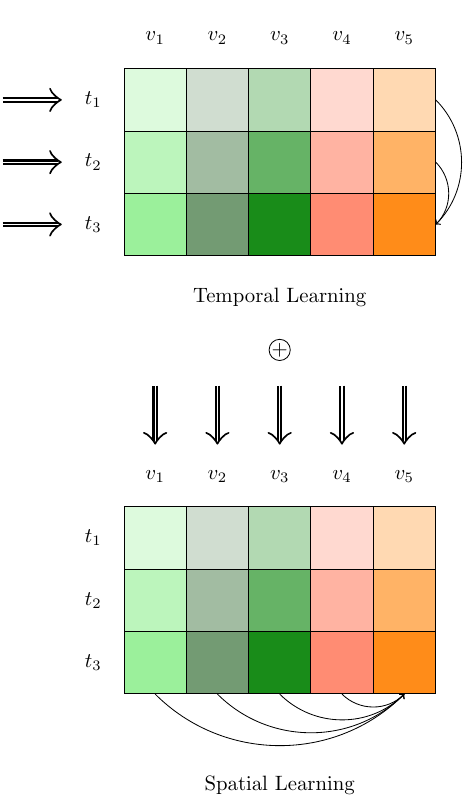}
        \caption{Existing Methods}
    \end{subfigure}
    ~
    \begin{subfigure}[b]{0.5\linewidth}
        \centering
        \includegraphics[width=\linewidth]{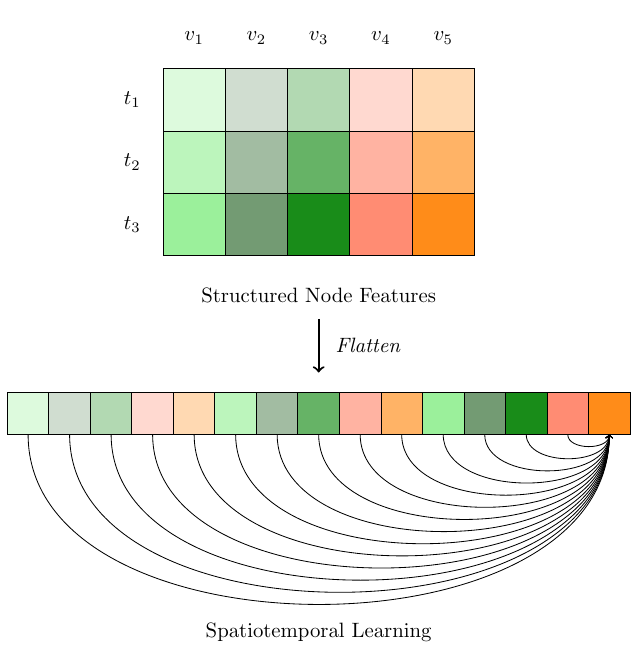}
        \caption{T-Graphormer}
    \end{subfigure}
    \caption{Information flow differences between existing methods and T-Graphormer. In part (a), the double arrows represent the cross-sectional input for each learning module, and the circle represents the fusion between the modules. For instance, GraphWaveNet has temporal modules (CNN) that treat the ``concatenated" node features as the hidden embeddings ($\bm{t_i}$), then its spatial learning modules (GNN) ``concatenate" the time features to produce node embeddings ($\bm{v_i}$). Such information flow is similar in attention-based models (e.g. STAEformer, PDFormer). The single arrows demonstrate that vector $\bm{t_3}$ is updated based on $\bm{t_1}\text{-}\bm{t_2}$, while $\bm{v_5}$ is updated based on $\bm{v_1}\text{-}\bm{v_4}$. In comparison, in part (b), each token in T-Graphormer attends to all other tokens.}
    \label{fig:big_picture}
\end{figure}

Most of the models we have discussed so far interleave temporal learning with spatial learning, transferring features between the time domain and the graph domain (Figure~\ref{fig:big_picture}). 
In another direction, \ac{STSGCN} \citep{song2020spatial} simultaneously captures spatial and temporal dependencies using a novel synchronous graph convolution operation. This is done by adding positional embeddings and connecting the nodes across time steps. While this approach avoids separate spatial and temporal learning modules, it is important to note that STSGCN treats the input as a graph and uses GCN for message passing. In comparison, T-Graphormer processes the input as a sequence. This underscores a fundamental difference between the two models.

\citet{grigsby2021long} also introduced Spacetimeformer, which flattens the structured data into a sequence before using the Transformer encoder. Our proposed method aligns their idea of using minimal architectural changes and data reshaping. However, Spacetimeformer fails to utilize structural information; it only uses two separate positional embeddings to index the time and space of each token in the original graph sequence. As shown in \citep{ying2021transformers}, connectivity data is crucial in Transformer-based graph learning.
\subsection{Transformers}
Transformers \citep{vaswani2017attention} were originally designed for \ac{NLP} applications, addressing limitations of earlier models like \ac{RNN}s \citep{rumelhart1986learning} and \ac{LSTM}s \citep{hochreiter1997long}, which struggled with long-range dependencies and training issues such as vanishing and exploding gradients \citep{pascanu2013difficulty}. In \ac{RNN}-based models, information from previous time steps is stored in a hidden state, requiring computational effort proportional to the distance between signals to relate them effectively. Transformers mitigate this challenge with an attention mechanism that processes the entire sequence as input, enabling tokens to interact directly in a constant number of operations. 

The multi-head attention mechanism empowers Transformers to model long-range dependencies and extract meaningful features, making them versatile for various data types, including text \citep{brown2020language, touvron2023llama, ouyang2022training}, images \citep{dosovitskiy2020image}, and graphs \citep{ying2021transformers, dwivedi2020generalization, kreuzer2021rethinking}. However, because Transformers are inherently designed for linear sequences, modifications are necessary. 
In 2020 and 2021, \citet{ying2021transformers, dwivedi2020generalization} proposed ways to apply Transformers directly on Graphs. Graphormer \citep{ying2021transformers} uses three types of structural encoding methods to account for the information lost during flattening. Centrality encoding is indexed by the in/out-degree of each node, signalling to the model the node importance. Spatial encoding is indexed by the connectivity between the nodes in the graph (e.g. shortest path). This structural encoding is directly added to the attention score as a bias term, such that the model can selectively pay more attention to nodes closer in space. Finally, edge encoding stores information about edge feature connectivity. This is also implemented as a bias term for the attention score.

\section{Preliminaries}
We introduce some notations and definitions used in traffic prediction, along with an overview of the self-attention mechanism in Transformers, which is used in T-Graphormer.
\subsection{Definitions}
Formally, a time series is a set of observations $X_t$, where $t$ denotes the time of observation. Forecasting is the task where given $T'$ historical data at time $t$, $\left(X_{t-T'+1}, X_{t-T'+2}, \dots, X_{t}\right)$, we wish to predict the next $T$ observations in the future $\left(X_{t+1}, X_{t+2}, \dots, X_{t+T}\right)$. 

In multivariate time series, each observation is a vector (or a matrix). Let $X_{t,i} \in \mathbb{R}^{1\times C}$ be the $i$th component of such observation at time $t$, and let
\begin{align*}
    \mathbf{X}_t &= (X_{t,1}^\top, X_{t,2}^\top, \dots, X_{t, N}^\top)^\top \in\mathbb{R}^{N \times C} \\
    \intertext{denote the observation at time $t$ with $N$ components each having $C$ features. The goal of multivariate time series forecasting is to predict}
    \bm{\mathcal{Y}} &= \left(\mathbf{X}_{t+1}^\top, \mathbf{X}_{t+2}^\top, \dots, \mathbf{X}_{t+T}^\top
\right)^\top \in \mathbb{R}^{T\times N \times C} \\
    \intertext{based on the historical data at time $t$}
    \bm{\mathcal{X}} &=\left(\mathbf{X}_{t-T'+1}^\top, \mathbf{X}_{t-T'+2}^\top, \dots, \mathbf{X}_{t}^\top\right)^\top \in \mathbb{R}^{T' \times N \times C}\\
\end{align*}
In our traffic prediction problem, the components of an observation lie in a graph $\mathcal{G} = (\mathcal{V}, \mathcal{E}, \mathbf{W})$. Here, $\mathcal{G}$ denotes a graph with the set of vertices $\mathcal{V}$ with size $N$ and the set of edges $\mathcal{E}$ with size $M$, and $\mathbf{W}\in \mathbb{R}^{N\times N}$ denotes the weighted adjacency matrix with $\mathbf{W}_{i,j}$ being the edge length. In other words, each $i$th component of the observation $\mathbf{X}_t$ at time $t$ is the node value of node $v_i$.

\subsection{Transformers}
Transformer architecture is a key component in many deep-learning models. It consists of multiple layers, each composed of two main parts: a self-attention mechanism and a position-wise \ac{FFN}.

Given an input sequence 
$H = \left( h_1^\top, \dots, h_l^\top \right)^\top \in \mathbb{R}^{l \times d}$ where $d$ is the hidden dimension and $h_i \in \mathbb{R}^{1 \times d}$ is the hidden representation at position $i$, three matrices are used to project $H$ to obtain $Q, K, V$. 
\begin{equation}\label{eq: QKV}
    Q = HW_Q, \quad K = HW_K, \quad V = HW_V
\end{equation}
where the three matrices are $W_Q \in \mathbb{R}^{d \times d_K}, W_K \in \mathbb{R}^{d \times d_K}, W_V \in \mathbb{R}^{d \times d_V}$ respectively. The similarity matrix $A$ is calculated as 
\begin{equation}\label{eq: attention_score}
    A = \frac{QK^\top}{\sqrt{d_K}},
\end{equation}
which measures the semantic similarity between query vectors in $Q$ and key vectors in $K$. Finally, this attention score matrix is used to retrieve the learned value vectors:
\begin{equation}\label{eq: attention}
    \mathrm{Attention}\left(H\right) = \mathrm{softmax}\left(
    A
    \right)V
\end{equation}
This defines one attention head. In multi-headed attention, the outputs of each head are concatenated and projected again. After the self-attention operation, each position $i$ in the sequence $H$ is processed independently and identically by a position-wise \ac{FFN}. The \ac{FFN} consists of two linear transformations with an activation function in between:
\begin{equation}\label{eq: FFN}
    \mathrm{FFN}(h_i) = \mathrm{activation}(h_i W_1 + b_1) W_2 + b_2.
\end{equation}

\section{T-Graphormer}
We first discuss how Graphormer can be easily extended in the temporal dimension to produce T-Graphormer, which models spatiotemporal data. We then discuss empirically beneficial implementation details.
\begin{figure}[ht]
    \centering
    \includegraphics[width=0.75\linewidth]{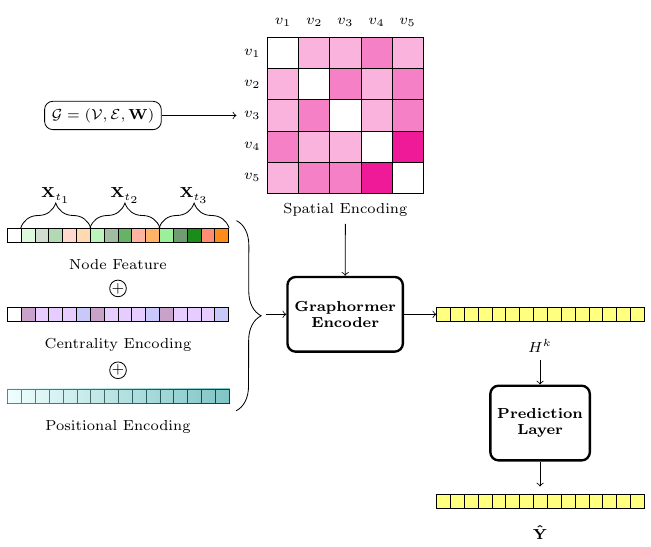}
    \caption{T-Graphormer model architecture. Centrality encoding and positional encoding are added to the node feature vector, which is then passed into the Graphormer encoder blocks. Edge weights are used to compute spatial encodings that determine the attention bias. Finally, the prediction layer maps the learned representation onto the output space. A \texttt{cls} token (white fill) is added to the beginning of the sequence. The illustration is consistent with the example in Figure ~\ref{fig:time_series}.}
    \label{fig:arch}
\end{figure}

\subsection{Node Feature Embedding}
Following Graph WaveNet by \citet{wu2019graph}, we project the observed node values from $C$ to $d$ dimensions with a linear layer 
$W_0\in\mathbb{R}^{C\times d}$. Let $\bm{\mathcal{X}}\in\mathbb{R}^{T' \times N \times C}$ be the historical data, we denote the initial node features as $\mathcal{X}\in\mathbb{R}^{T' \times N \times d}$.

\subsection{Structural Encodings}
As discussed in section ~\ref{sec: related_works}, structured data must be flattened before it can be processed by Transformers. We vectorize $\mathcal{X}$ into 
$$
\mathbf{x} = (X_{t-T'+1,1}, \dots, X_{t+2,N}, \dots,   X_{t,N})\in\mathbb{R}^{l \times d}
$$
where $l = T'\times N$.
However, information is lost during this process, so when applying Transformers to spatiotemporal data, structural encoding methods are essential to introduce inductive biases that inform the model about the structure in $\mathcal{X}$. In \ac{T-Graphormer}, we extend the structural encoding techniques introduced in Graphormer \citep{ying2021transformers} to effectively capture the spatiotemporal relationships within the data. This is mainly done in two ways: modifications to the node features and modifications to the attention mechanism (see Figure ~\ref{fig:arch}).

Let $x_{t,i} \in \mathbb{R}^{1\times d}$ be the feature vector of the $i$th node at time $t$, centrality encoding is added to $x_{t,i}$ to inform the model about node importance. Intuitively, in a traffic network, intersections with more road connections have significant effect on the downstream traffic conditions. Let $\mathrm{deg}^-(v)$ denote the in-degree of node $v$, and $\mathrm{deg}^+(v)$ denote the out-degree, centrality encoding, $Z^-\in\mathbb{R}^{\mathrm{deg}^-(V)\times d}$ and $Z^+ \in\mathbb{R}^{\mathrm{deg}^+(V)\times d}$, are real-valued learnable embedding matrices indexed by $\mathrm{deg}(v)$ where the number of rows corresponds to the maximum node in-degree and out-degree in $\mathcal{G}$ respectively. Concretely, centrality encoding is applied to all initial node features before entering the first Transformer block
\begin{equation}\label{eq: centrality}
    h_{t,i}^0 = x_{t,i} + z_{\mathrm{deg}^-(v_i)}^- + z_{\mathrm{deg}^+(v_i)}^+.
\end{equation}
Note that the encoding is time-agnostic and is determined solely by the degree of the corresponding node. In other words, for $x_{t_1, i}$ and $x_{t_2,i}$ the centrality encodings are both $z_\mathrm{deg}(v_i)$ (as shown in Figure ~\ref{fig:arch} centrality encoding). Also in cases of undirected graphs, only $Z \in\mathbb{R}^{\mathrm{deg}(V)\times d}$ is used. 

To account for the temporal structure, we also add learnable positional encoding \citep{vaswani2017attention, dosovitskiy2020image} to the initial node features. Let $P\in \mathbb{R}^{l\times d}$ denote the learned matrix where $l = T'N$, we update the initial node features in equation ~\ref{eq: centrality} as
\begin{equation}\label{eq: positional}
    h_{t,i}^0 = x_{t,i} + z_{\mathrm{deg}^-(v_i)}^- + z_{\mathrm{deg}^+(v_i)} + p_{t,i}.
\end{equation}
Note that, unlike centrality encoding, positional encoding vectors $p_{t,i}$ are specific to each token in the vectorized time series. While \citet{feichtenhofer2022masked} adopts separate positional embeddings when applying \ac{ViT} on videos, with one for time and one for space, we find T-Graphormer performs better when positional embedding is learned together. 

While the attention mechanism in Transformers has an effective global receptive field, the spatial encoding method is another important structural bias that improves spatial learning. Given the graph network $\mathcal{G} = (\mathcal{V}, \mathcal{E}, \mathbf{W})$, the model can leverage the weighted adjacency matrix to determine which nodes are closer in topological space for updating representations. Let $\phi (v_i, v_j): \mathcal{V} \times \mathcal{V} \mapsto \mathbb{R}$ denote a function that measures the spatial relation between nodes, the attention score (equation ~\ref{eq: attention_score}) between tokens $(t_1, i)$ and $(t_2, j)$ is updated as:
\begin{equation}\label{eq: spatial_encoding}
    A_{(t_1, i), (t_2, j)} = 
    \frac{\left(h_{t_1, i}W_Q\right)\left(h_{t_2, i}W_K\right)^\top}
    {\sqrt{d}} + b_{\phi(i,j)}
\end{equation}
where $b_{\phi(i,j)}$ is the learnable attention bias scalar indexed by $\phi(i,j)$ from $\Phi$. 

Similar to Graphormer, we define $\phi$ as the shortest path distance (SPD) between nodes and $\Phi \in \mathbb{R}^{N\times N}$ as the matrix storing the SPD values for all node pairs. For example, if the longest path between any two nodes in $\mathcal{G}$ is 3, the learnable attention bias embedding $B$ will have shape $3 \times 3 \times \textit{number of heads}$. Like centrality encoding, spatial encoding is time-independent; for nodes $i$ and $j$, the term remains consistent across time steps (see Figure ~\ref{fig:arch}, Spatial Encoding).

\subsection{Implementation Details}\label{sec: implement}
To project the learned node representations from the encoder to the output space, we use two linear layers to project the hidden representation at the last layer $H_k\in\mathbb{R}^{l\times d}$ from $d$ dimensions to $\frac{d}{2}$, and finally to $C$.

Before passing the node features through the Transformer blocks, we experiment with adding special tokens to the sequence:
\begin{itemize}
    \item No special token added, the length of the sequence is $T'\times N$.
    
    \item Adding a \texttt{cls} token to the start of the sequence (see Figure ~\ref{fig:arch}), resulting in $(T'\times N) + 1$ tokens. Similar to the virtual node in Graphormer, a learnable embedding of size $\mathbb{R}^{1\times d}$ is introduced and concatenated with the initial node features $\mathcal{X}$. This token acts as a supernode, aggregating information from the entire sequence and propagating it back.
    
    \item Adding a \texttt{graph} token to the start of each graph signal, resulting in $T'\times (N+1)$ tokens. Similar to the \texttt{cls} token, a learnable embedding is used. Its function is analogous to the \texttt{sep} token in \ac{BERT} \citep{devlin2018bert}, serving as a delimiter for graph signals at different time steps. 
\end{itemize}
Virtual connections are formed between special tokens and other nodes, and $b_{\phi(\texttt{token}, v)}$ is learned separately.

We find that using layer normalization \citep{ba2016layer} before multi-headed attention and \ac{FFN} improves performance. We also find \ac{GELU} to be the best activation function.

\section{Experiments}
In this section, we detail the experimental settings used to evaluate T-Graphormer on two traffic prediction datasets. We compare T-Graphormer's performance against 10 baseline methods. Additionally, to gain deeper insights into T-Graphormer, we analyze its attention scores and conduct ablation studies.

\subsection{Datasets}\label{sec: datasets}
Following \citet{li2018diffusion, wu2019graph, wu2020connecting, shao2022pre}, we focus on evaluating \ac{T-Graphormer} on two commonly used spaiotemporal forecasting datasets (see Table ~\ref{tab:dataset} for details):
\begin{itemize}
    \item PEMS-BAY \citep{chen2001freeway}: A traffic speed dataset collected from California Transportation Agencies (CalTrans) Performance Measurement System (PeMS).
    \item METR-LA \citep{jagadish2014big}: A traffic speed dataset collected from loop detectors on the highway of Los Angeles County.
\end{itemize}
For both datasets, we follow the same pre-processing implementation as \citet{li2018diffusion}\footnote{https://github.com/liyaguang/DCRNN}. Each sample $\mathbf{X}_t$ is a 5-minute traffic speed reading from all sensors in the network. We aggregate 12 consecutive samples (representing a 1-hour context window) to construct $\bm{\mathcal{X}}\in\mathbb{R}^{12\times N\times 1}$. The ground truth $\bm{\mathcal{Y}}$ consists of the next $T$ future traffic speed readings from all sensors. To ensure consistency with the baseline methods, we evaluate the models on predictions for $T=\{3, 6, 12\}$, corresponding to 15 minutes, 30 minutes, and 1 hour. We also concatenate a one-hot-encoding vector signifying the time of day to all node values at time $t$.

The data is split such that approximately 70\% is used for training, 20\% for testing, and 10\% for validation. To avoid data leakage in the traffic prediction task, the time slices are kept in their original order before splitting. Only the training dataset is shuffled for different training iterations and used to perform $Z$-score normalization on the validation and testing datasets. To construct the graph $\mathcal{G} = (\mathcal{V}, \mathcal{E}, \mathbf{W})$, we follow the same implementation as \citet{li2018diffusion}.

\subsection{Settings}\label{sec: settings}
We conduct extensive experiments to investigate how different hyperparameters (e.g. learning rate, gradient clipping, layer decay) affect the performance. Three different Transformer configurations were evaluated with $d=\{64, 128, 192\}$, $k=\{6, 6, 8\}$, and $\text{number of heads} = \{2, 4, 6\}$. 

T-Graphormer is trained by minimizing the Huber loss \citep{huber1992robust} ($\delta=1.5$) between predicted values $\mathbf{\hat{Y}}$ and ground truth $\bm{\mathcal{Y}}$ for $T=12$ . When evaluating the prediction performance on horizons 3 and 6, we simply remove extra token values. 
After training, model configurations with the lowest mean absolute error (MAE) on the validation dataset are selected for testing. T-Graphormer is implemented in \texttt{PyTorch} \citep{paszke2019pytorch} and utilizes distributed data parallelism to speed up training and increase batch size.

We compared against the following baseline methods:
\begin{enumerate*}
\item Vector Autoregressive (VAR) \citep{zivot2006vector},
\item Support Vector Regression (SVR) \citep{smola2004tutorial},
\item FC-LSTM \citep{sutskever2014sequence},
\item Graph Multi-Attention Network (GMAN) \citep{zheng2020gman},
\item \ac{STEP} \citep{shao2022pre},
\item PDFormer \citep{jiang2023pdformer},
\item STAEformer \citep{liu2023spatio}
\end{enumerate*}.

For more information on experimental settings, see Appendix at section ~\ref{sec: appendix}.

\subsection{Main Results}
\begin{table*}[htb]
    \centering
    \resizebox{0.9\linewidth}{!}{%
    \begin{tabular}{llccccccccc}
        \toprule
        \multirow{2}{*}{\textbf{Dataset}} & \multirow{2}{*}{\textbf{Method}} & \multicolumn{3}{c}{\textbf{Horizon 3}} & \multicolumn{3}{c}{\textbf{Horizon 6}} & \multicolumn{3}{c}{\textbf{Horizon 12}} \\ 
        \cmidrule(lr){3-5}\cmidrule(lr){6-8}\cmidrule(lr){9-11}
        & & MAE & RMSE & MAPE & MAE & RMSE & MAPE & MAE & RMSE & MAPE \\
        \midrule
        
        \multirow{11}{*}{PEMS-BAY} 
        & VAR & 1.74 & 3.16 & 3.60 & 2.32 & 4.25 & 5.00 & 2.93 & 5.44 & 6.50 \\
        & FC-LSTM & 2.05 & 4.19 & 4.80 & 2.20 & 4.55 & 5.20 & 2.37 & 4.96 & 5.70 \\
        & DCRNN & 1.38 & 2.95 & 2.90 & 1.74 & 3.97 & 3.90 & 2.07 & 4.74 & 4.90 \\
        & STGCN & 1.36 & 2.96 & 2.90 & 1.81 & 4.27 & 4.17 & 2.49 & 5.69 & 5.79 \\
        & Graph WaveNet & 1.30 & 2.74 & 2.73 & 1.63 & 3.70 & 3.67 & 1.95 & 4.54 & 4.63 \\
        & ASTGCN & 1.52 & 3.13 & 3.22 & 2.01 & 4.27 & 4.48 & 2.61 & 5.42 & 6.00 \\
        & GMAN & 1.34 & 2.91 & 2.86 & 1.63 & 3.76 & 3.68 & 1.86 & 4.32 & 4.37 \\
        & PDFormer & 1.32 & 2.83 & 2.78 & 1.64 & 3.79 & 3.71 & 1.91 & 4.43 & 4.51 \\
        & STAEformer & 1.31 & 2.78 & 2.76 & 1.62 & 3.68 & 3.62 & 1.88 & 4.34 & 4.41 \\
        & STEP & \underline{1.26} & \underline{2.73} & \underline{2.59} & \underline{1.55} & \underline{3.58} & \underline{3.43} & \underline{1.79} & \underline{4.20} & \underline{4.18} \\
        \cmidrule(lr){2-11}
        & \textbf{T-Graphormer} & \textbf{1.16} & \textbf{2.10} & \textbf{2.38} & \textbf{1.38} & \textbf{2.64} & \textbf{2.97} & \textbf{1.63} & \textbf{3.20} & \textbf{3.65} \\
        \midrule\midrule
        
        \multirow{11}{*}{METR-LA} 
        & VAR & 4.42 & 7.80 & 13.00 & 5.41 & 9.13 & 12.70 & 6.52 & 10.11 & 15.80 \\
        & FC-LSTM & 3.44 & 6.30 & 9.60 & 3.77 & 7.23 & 10.09 & 4.37 & 8.69 & 14.00 \\
        & DCRNN & 2.77 & 5.38 & 7.30 & 3.15 & 6.45 & 8.80 & 3.60 & 7.60 & 10.50 \\
        & STGCN & 2.88 & 5.74 & 7.62 & 3.47 & 7.24 & 9.57 & 4.59 & 9.40 & 12.70 \\
        & Graph WaveNet & 2.69 & 5.15 & 6.90 & 3.07 & 6.22 & 8.37 & 3.53 & 7.37 & 10.01 \\
        & ASTGCN & 4.86 & 9.27 & 9.21 & 5.43 & 10.61 & 10.13 & 6.51 & 12.52 & 11.64 \\
        & GMAN & 2.80 & 5.55 & 7.41 & 3.12 & 6.49 & 8.73 & 3.44 & 7.35 & 10.07 \\
        & PDFormer & 2.83 & 5.45 & 7.77 & 3.20 & 6.46 & 9.19 & 3.62 & 7.47 & 10.91 \\
        & STAEformer & 2.65 & 5.11 & 6.85 & 2.97 & 6.00 & 8.13 & \underline{3.34} & 7.02 & 9.70 \\
        & STEP & \textbf{2.61} & 4.98 & \textbf{6.60} & \underline{2.96} & \underline{5.97} & \underline{7.96} & 3.37 & \underline{6.99} & \underline{9.61} \\
        \cmidrule(lr){2-11}
        & \textbf{T-Graphormer} & \underline{2.62} & \textbf{4.93} & \underline{6.61} & \textbf{2.90} & \textbf{5.53} & \textbf{7.56} & \textbf{3.19} & \textbf{6.12} & \textbf{8.62} \\
        \bottomrule
    \end{tabular}}
    \caption{Spatiotemporal forecasting results on two traffic network datasets. The best performance for each metric is in bold, and the second-best is underlined. MAPE is reported in percentage.}
    \label{tab:pred_results}
\end{table*}

T-Graphormer demonstrates exceptional performance in traffic prediction. As shown in Table ~\ref{tab:pred_results}, it outperforms all other models across every metric when predicting the next 6 and 12 time steps (30-minute and 1-hour windows). Compared to state-of-the-art methods, it reduces RMSE by \textbf{24\%} and \textbf{12\%} on the PEMS-BAY and METR-LA datasets. For MAPE, it reduces the error by \textbf{13\%} and \textbf{10\%}, respectively. 

\begin{figure}[htb]
    \centering
    \begin{subfigure}[b]{0.4\textwidth}
        \centering
        \includegraphics[width=\textwidth]{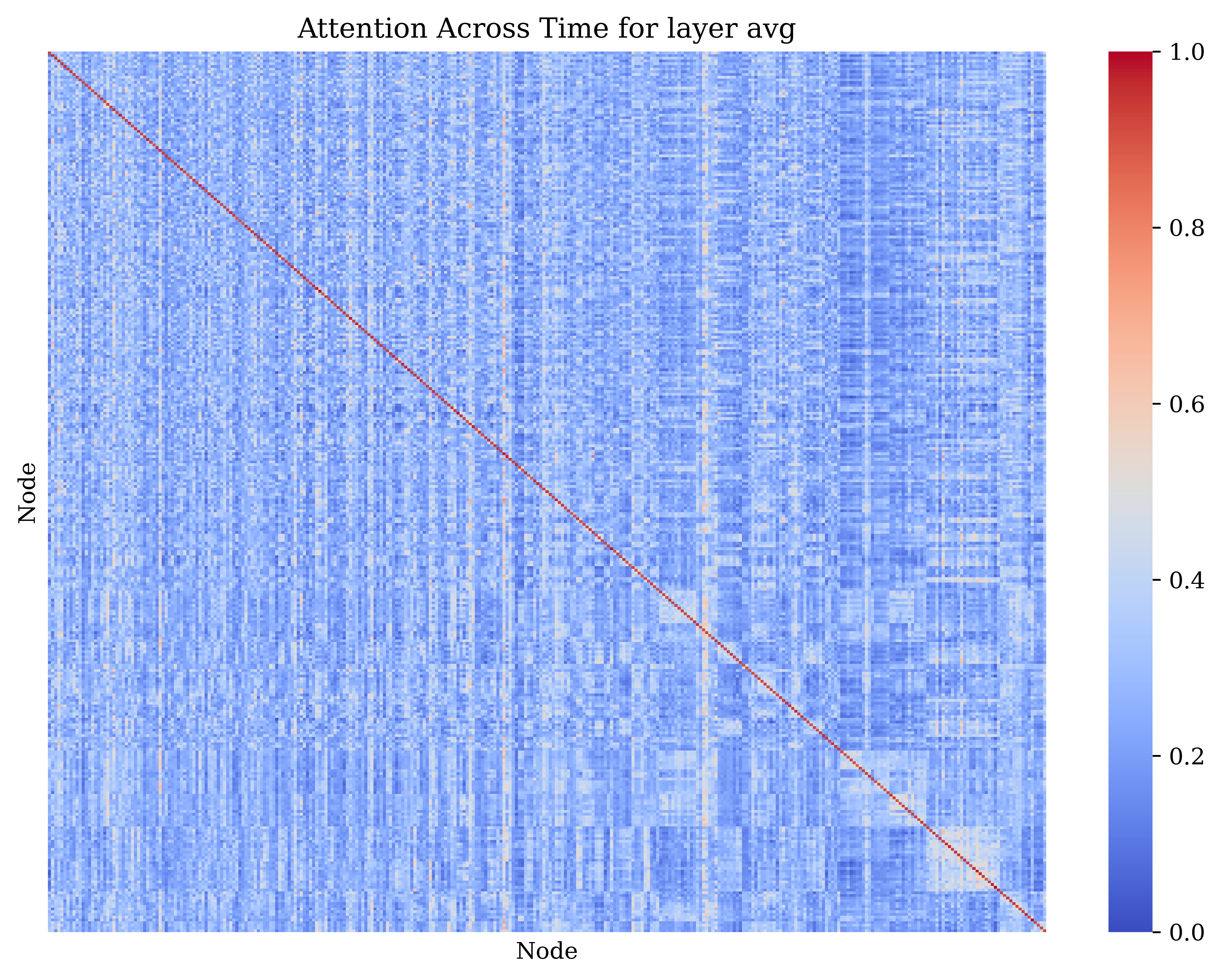}
        \caption{Node-node attention scores}
    \end{subfigure}
    ~
    \begin{subfigure}[b]{0.4\textwidth}
        \centering
        \includegraphics[width=\textwidth]{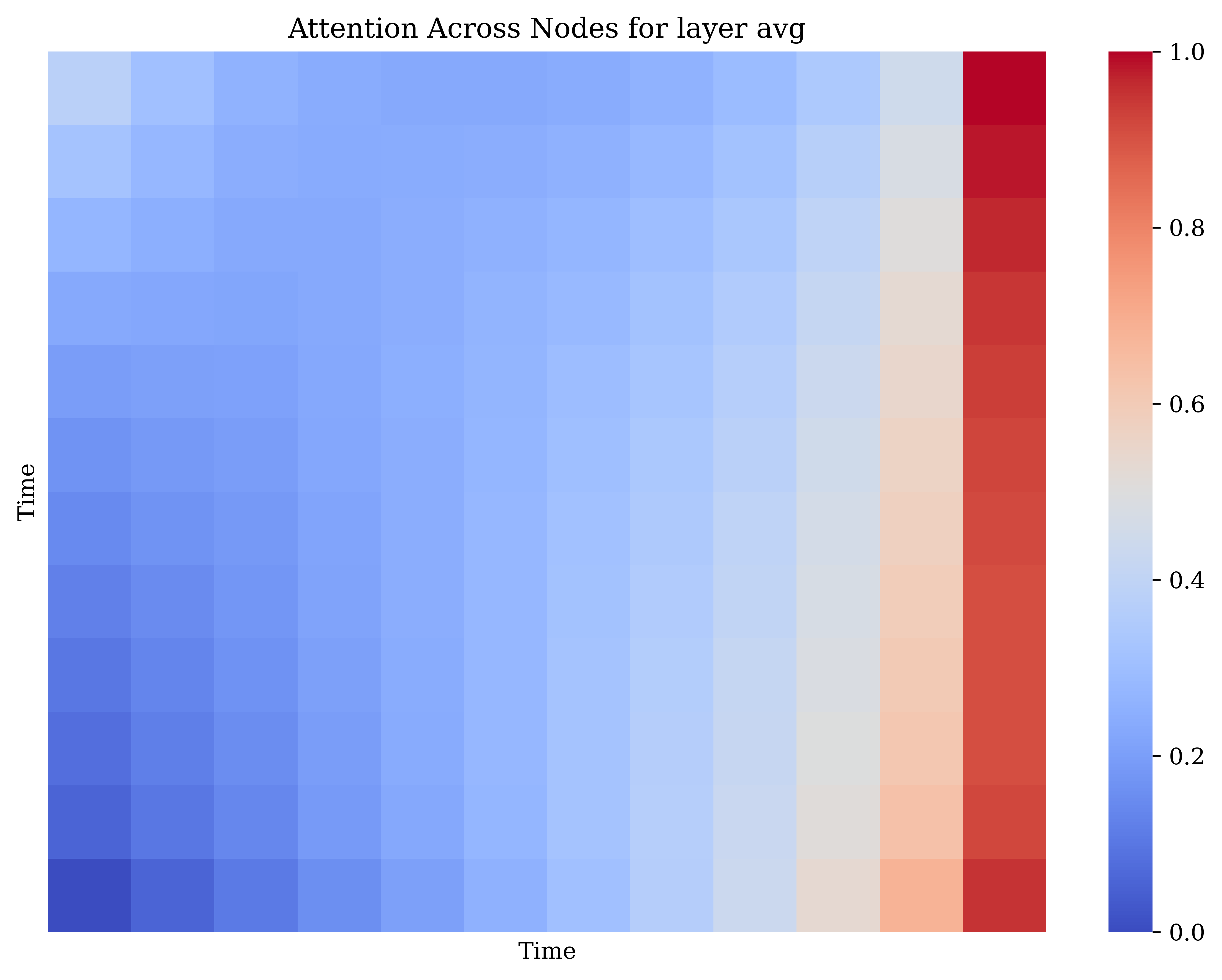}
        \caption{Time-time attention scores}
    \end{subfigure}    
    \caption{Attention heatmap averaged across time and space shows T-Graphormer learns spatial and temporal relationships. On the PEMS-BAY dataset, node-node attention scores reveal strong self-attention, symmetric clusters in the bottom right corner, and selectivity to important nodes (shown in columns with 0.5 values). Time-time attention scores reveal strong emphases on recent time points. It also shows temporal grouping, where earlier and later time points attend within each other but not between (shown by the colour gradient in the top-left and bottom-right corners). This pattern is more obvious in the METR-LA dataset in Figure ~\ref{fig:metr-attn}.}
    \label{fig:pems-attn}
\end{figure}

We attribute this success to the minimal structural biases in T-Graphormer. It has been observed in images and videos \citep{dosovitskiy2020image, he2022masked, feichtenhofer2022masked} that when models (Transformers) break free from the traditional information flow imposed by data structure, predefined relationships are learned directly from the data, subsequently improving task performance. Here, we show that this trend also applies to traffic data. In Figure~\ref{fig:pems-attn}, we see T-Graphormer learns to selectively attend to important nodes and time points without explicit spatial and temporal dimensions in model architecture. Note that to obtain these heatmaps, we average the attention score between tokens ($T'N \times T'N$) across samples, then layers, and finally time or node depending on the dimension of interest. Thus, these heatmaps are cross-sectional, and the results combined suggest that T-Graphormer also learns spatiotemporal dependencies.


We also observe that T-Graphormer performed differently across the two datasets as the prediction task on METR-LA is harder. In Figure ~\ref{fig:val_plot}, the validation MAPE is around 4 for PEMS-BAY but increased to around 9 for METR-LA, and in Figure~\ref{fig:loss_plot}, training loss of the best model is much lower in PEMS-BAY (1.44) than in METR-LA (2.55). This is unsurprising as the traffic speed standard deviation of METR-LA (19.49) is much higher than that of PEMS-BAY (9.44).

We also notice that in Figure ~\ref{fig:loss_plot}, the training and validation loss curves follow much more closely in PEMS-BAY than in METR-LA, a sign that the model is overfitting on METR-LA (although this is not observed in other validation metrics). This can be explained through the fact that PEMS-BAY has 50\% more samples than METR-LA (52116 vs.\ 34727). We inspect this behaviour from another perspective in Figure ~\ref{fig:train_val_diff_plot}. Between the three model sizes, a trend emerges towards the end of training, where bigger models have higher loss differences. This is consistent with the recent findings on the empirical scaling laws for training LLM \citep{kaplan2020scaling}. Using equation (6.6) from this work: $D\propto N^{0.74}$, we find that the datasets used in the experiment are only optimal for training the mini models. We confirm this theoretical result in Table~\ref{tab:add_pred_results}.

\subsection{Ablation Experiments}
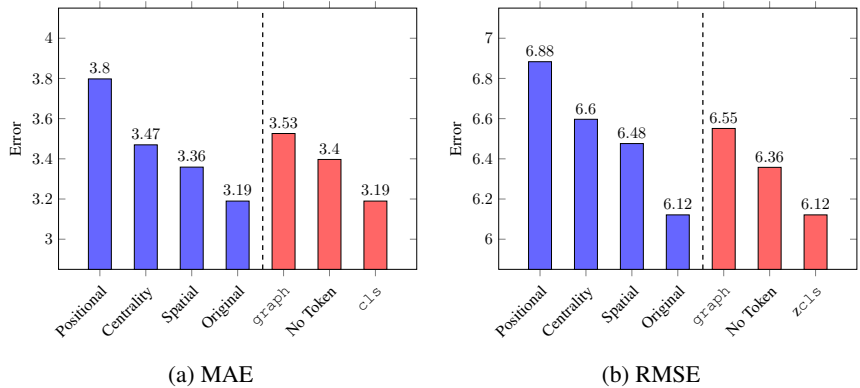
\begin{figure}[ht]
    \centering
    \begin{subfigure}[b]{0.4\textwidth}
        \centering
        \resizebox{\linewidth}{!}{
        \begin{tikzpicture}
        \begin{axis}[
            x=1cm,
            ybar,
            bar width=0.5cm,
            symbolic x coords={pos, Centrality, Spatial, Original, graph, No Token, cls},
            xtick={pos, Centrality, Spatial, Original, graph, No Token, cls},
            xticklabels={{Positional}, {Centrality}, {Spatial}, {Original}, {\texttt{graph}}, {No Token}, {\texttt{cls}}},
            xticklabel style={rotate=45, anchor=north east},
            ymin=3, ymax=4,
            ylabel={Error},
            nodes near coords,
            enlargelimits=0.15
        ]
        \addplot[bar shift=0pt, fill=blue!60]  coordinates {(pos, 3.79762) (Centrality, 3.46956) (Spatial, 3.35875) (Original, 3.18976)};
        \addplot[bar shift=0pt, fill=red!60]  coordinates {(graph, 3.52599) (No Token, 3.39689) (cls, 3.18976)};
        
        \draw [dashed, thick] ({rel axis cs:0.57,0}) -- ({rel axis cs:0.57,1});
    \end{axis}
    \end{tikzpicture}
        }
        \caption{MAE}
    \end{subfigure}
    ~
    \begin{subfigure}[b]{0.4\textwidth}
        \centering
        \resizebox{\linewidth}{!}{
        \begin{tikzpicture}
        \begin{axis}[
            x=1cm,
            ybar,
            bar width=0.5cm,
            symbolic x coords={pos, Centrality, Spatial, Original, graph, No Token, cls},
            xtick={pos, Centrality, Spatial, Original, graph, No Token, cls},
            xticklabels={{Positional}, {Centrality}, {Spatial}, {Original}, {\texttt{graph}}, {No Token}, {z\texttt{cls}}},
            xticklabel style={rotate=45, anchor=north east},
            ymin=6, ymax=7,
            ylabel={Error},
            nodes near coords,
            enlargelimits=0.15
        ]
            \addplot[bar shift=0pt, fill=blue!60] coordinates {(pos, 6.88323) (Centrality, 6.59676) (Spatial, 6.47595) (Original, 6.12108)};
            \addplot[bar shift=0pt, fill=red!60] coordinates {(graph, 6.55111) (No Token, 6.3576) (cls, 6.12108)};
            
            \draw [dashed, thick] ({rel axis cs:0.57,0}) -- ({rel axis cs:0.57,1});
        \end{axis}
        \end{tikzpicture}
        }
        \caption{RMSE}
    \end{subfigure}
    \caption{Ablation results on the METR-LA dataset with a forecasting horizon of 12. Ablation trends are similar for MAPE (Figure~\ref{fig: ablation_supp}). Left: Structural encoding ablations. Right: Special token ablations.}
    \label{fig: ablation}
\end{figure}
In this section, we examine the efficacy of the added structural encoding methods and assess the impact of special tokens. This is done by re-training the best-performing model with the same configuration but with the missing components. 

It is evident from Figure ~\ref{fig: ablation} that positional encoding and spatial encoding are the critical structural and temporal inductive biases when applying Transformers to spatiotemporal data. When positional encoding is removed, the model fails to predict accurately, leading to a \textbf{19\%} increase in MAE. This is unsurprising since Transformers lack recurrence or convolutional mechanisms and thus rely on positional encoding for temporal indexing.

Similarly, T-Graphormer relies on spatial biases for traffic prediction. When centrality encoding is removed, the performance drops by \textbf{9\%}. When spatial encoding is removed, the performance drops by \textbf{5\%}. We find this result surprising, as it implies the model learns more from node degree than the distance between node pairs. Overall, our findings are consistent with the growing literature in traffic prediction \citep{li2018diffusion, yu2018spatio, wu2019graph, guo2019attention}, where adding structural biases improves prediction performance.

Consistent with the work by \citet{ying2021transformers}, we also observe that adding special tokens improves model performance. We also find that adding \texttt{cls} or \texttt{graph} token noticeably improves prediction performance, and \texttt{cls} outperforms \texttt{graph}.


\section{Conclusion}
We introduce a novel framework for modelling spatiotemporal data. By leveraging structural encoding methods and extending them along the temporal dimension, we show that the Transformer architecture should be directly applied to the data without separate learning steps on two data dimensions. This integration allows T-Graphormer to capture spatiotemporal dependencies simultaneously with minimal inductive bias. Building on this foundation, future work can readily incorporate domain knowledge into the architecture, such as using adaptive spatiotemporal embeddings from STAEformer or custom spatial attention mechanisms from PDFormer.

However, our study has limitations that merit further discussion. While T-Graphormer shows strength in modelling spatiotemporal data, it does so with high memory cost with $\mathcal{O}(n^2)$ complexity. Flattening the entire time series into a sequence significantly increases context length when each observation has hundreds of components. For example, in the PEMS-BAY dataset, adding one time step increases the context length by 325 (the number of nodes). This constrains T-Graphormer's applicability to datasets with large networks or longer time windows. For instance, training T-Graphormer on the large-scale traffic forecasting dataset LargeST \citep{liu2024largest}, which includes up to 8600 nodes, is impossible without architectural changes. One would need to incorporate techniques such as sparse attention \citep{beltagy2020longformer} to mitigate this issue.

Additionally, this work focuses on applying T-Graphormer to static graphs, where the graph structure remains constant over time. Dynamic graphs, where the graph structure evolves, present a promising area for extension. For instance, \citet{shang2021discrete} demonstrated that graph learning techniques can effectively reparameterize dynamic graphs for forecasting tasks. Adapting T-Graphormer to dynamic graphs would require strategies such as introducing a maximum graph size and applying padding to handle structural changes over time. These changes can extend T-Graphormer to a broader range of spatiotemporal forecasting tasks and provide deeper insights into its graph representation learning abilities. 
%


\section*{Acknowledgement}
H.Y.B. was supported by the Master's Research Scholarship (Bourses de maîtrise en recherche) from Fonds de recherche du Québec – Nature et technologies (FRQNT). We are also grateful to the Digital Research Alliance of Canada for providing access to the high-performance computing resources. Specifically, this research was enabled in part by software provided by the Digital Research Alliance of Canada \href{https://alliancecan.ca/}{\url{https://alliancecan.ca/}}.
\section*{Reproducibility Statement}
For reproducibility, we include links to the code repository, the datasets, and the pre-trained model weights. We also summarize the training configurations and environments. All experimental logs are saved and available online at Weights and Biases upon request.
\begin{itemize}
    \item Code: \url{https://github.com/rdh1115/T-Graphormer}
    \item Datasets \citep{li2018diffusion, song2020spatial}: \url{https://github.com/liyaguang/DCRNN},\url{https://github.com/Davidham3/STSGCN}
    \item Pre-trained model weights: \url{https://www.kaggle.com/models/markbai/t-graphormer_pred_mini}
\end{itemize}
\bibliography{references}
\bibliographystyle{iclr2025_conference}

\appendix
\section{Appendix}\label{sec: appendix}
\begin{figure}[ht]
    \centering
    \resizebox{0.5\linewidth}{!}{
    \begin{tikzpicture}
        \begin{axis}[
            x=1cm,
            ybar,
            bar width=0.5cm,
            symbolic x coords={pos, Centrality, Spatial, Original, graph, No Token, cls},
            xtick={pos, Centrality, Spatial, Original, graph, No Token, cls},
            xticklabels={{Positional}, {Centrality}, {Spatial}, {Original}, {\texttt{graph}}, {No Token}, {\texttt{cls}}},
            xticklabel style={rotate=45, anchor=north east},
            ymin=8, ymax=10,
            ylabel={MAPE},
            nodes near coords,
            enlargelimits=0.15
        ]
            \addplot[bar shift=0pt,fill=blue!60] coordinates {(pos, 9.7376) (Centrality, 8.658) (Spatial, 9.0443) (Original, 8.6228)};
            \addplot[bar shift=0pt,fill=red!60] coordinates {(graph, 8.9584) (No Token, 8.8708) (cls, 8.6228)};
            
            \draw [dashed, thick] ({rel axis cs:0.57,0}) -- ({rel axis cs:0.57,1});
        \end{axis}
        \end{tikzpicture}
    }
    \caption{Ablation results on the METR-LA dataset with a forecasting horizon of 12. Left: Structural encoding ablations. Right: Special token ablations.}
    \label{fig: ablation_supp}
\end{figure}

\subsection{Model Behaviour}
\subsubsection{Attention Analysis}
\begin{figure}[htb]
    \centering
    \begin{subfigure}[b]{0.45\textwidth}
        \centering
        \includegraphics[trim={0 0 2cm 0},clip,width=\textwidth]{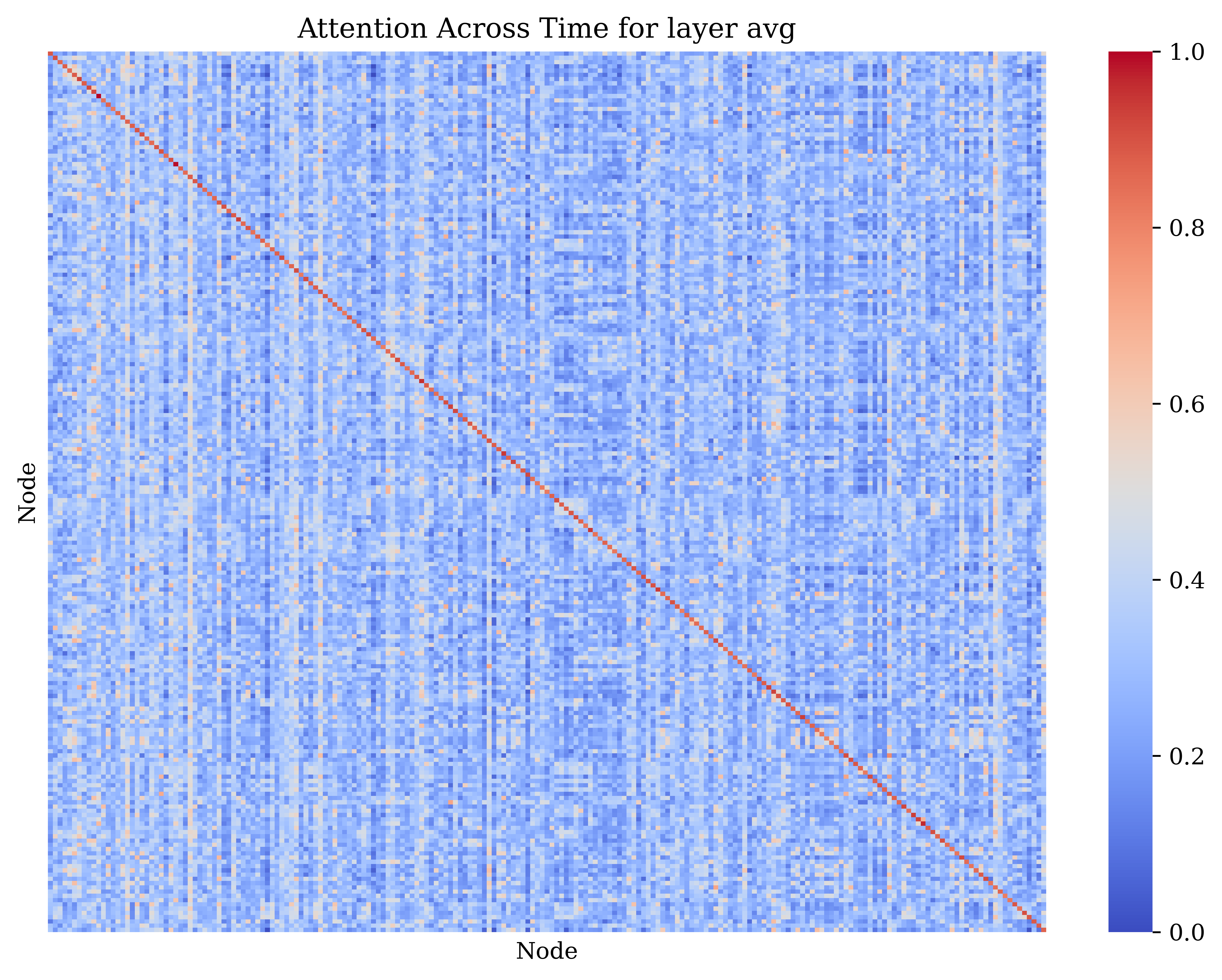}
        \caption{METR-LA node-node attention scores}
    \end{subfigure}
    ~
    \begin{subfigure}[b]{0.45\textwidth}
        \centering
        \includegraphics[trim={0 0 2cm 0},clip,width=\textwidth]{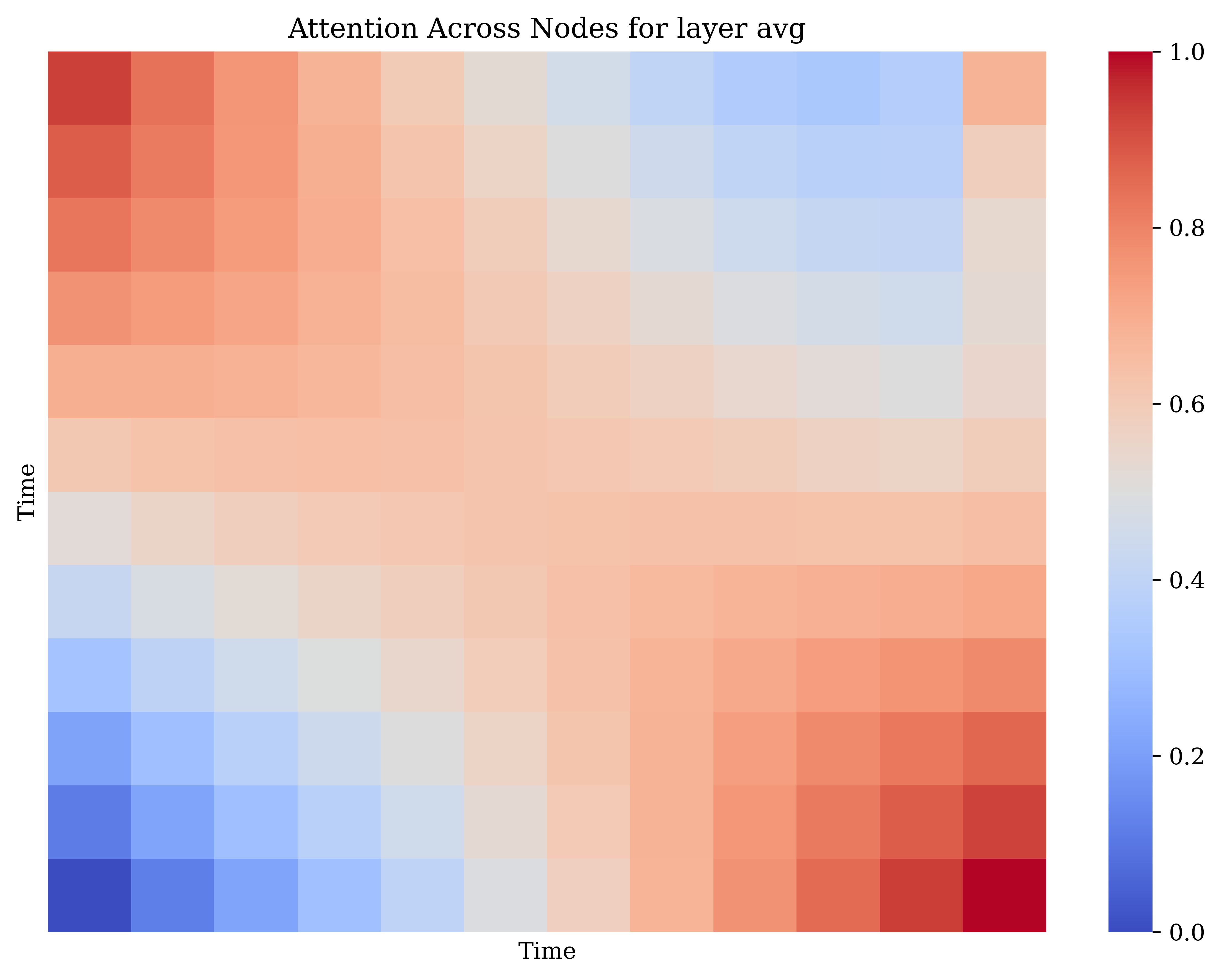}
        \caption{METR-LA time-time attention scores}
    \end{subfigure}
    \bigskip
    \begin{subfigure}[b]{0.45\textwidth}
        \centering
        \includegraphics[trim={0 0 2cm 0},clip,width=\textwidth]{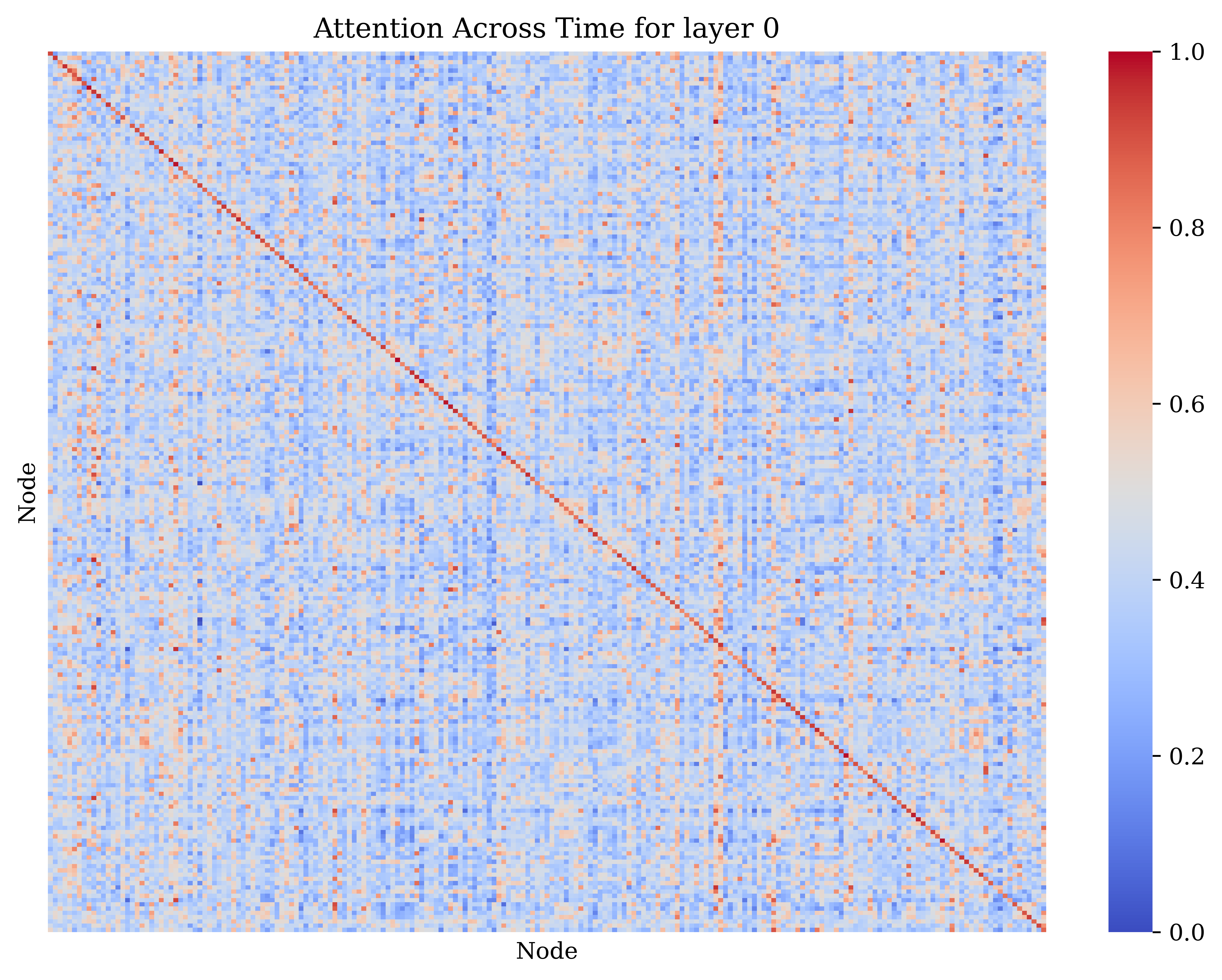}
        \caption{METR-LA node-node attention scores layer 1}
    \end{subfigure}
    ~
    \begin{subfigure}[b]{0.45\textwidth}
        \centering
        \includegraphics[trim={0 0 2cm 0},clip,width=\textwidth]{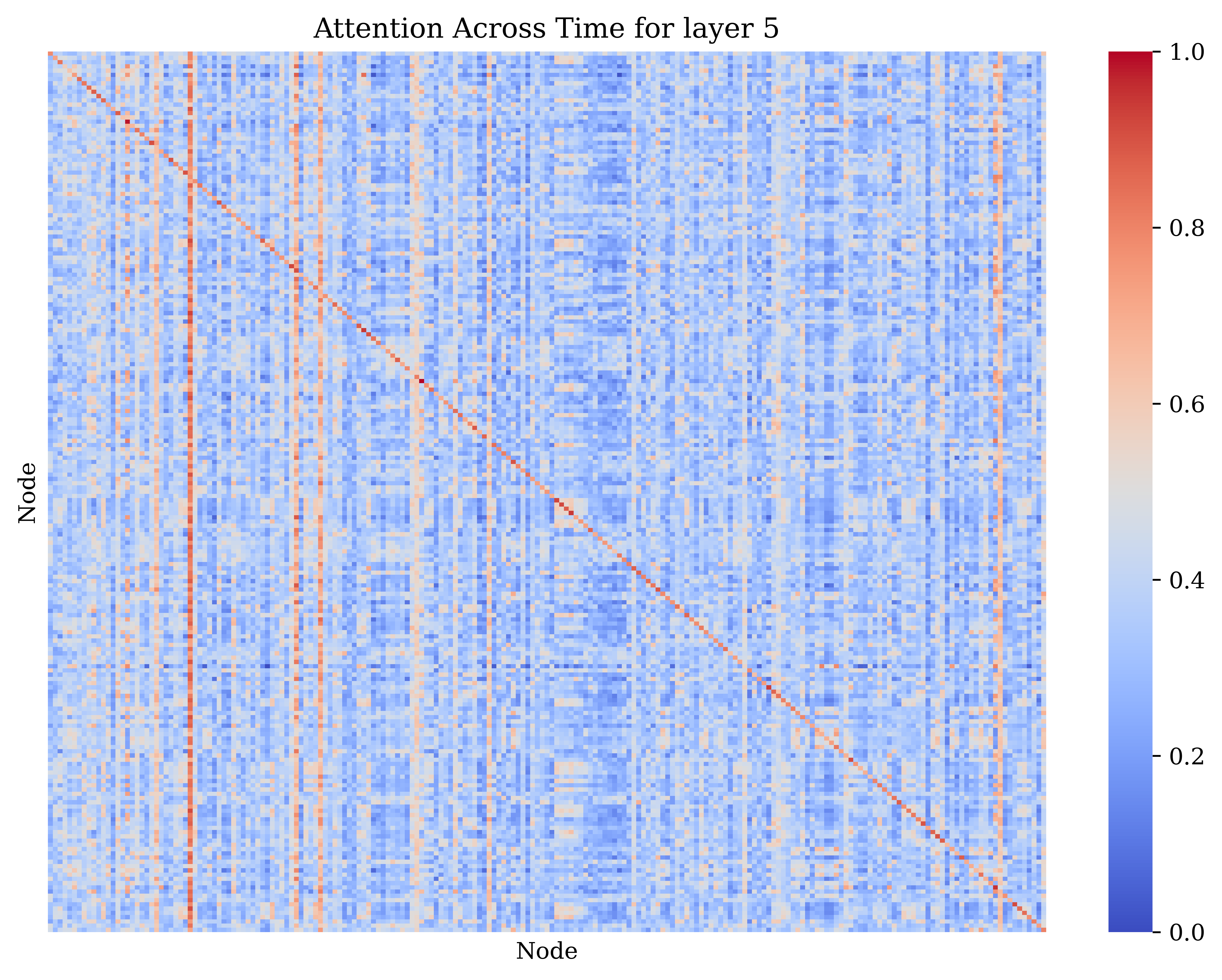}
        \caption{METR-LA node-node attention scores layer 6}
    \end{subfigure}
    \bigskip
    \begin{subfigure}[b]{0.45\textwidth}
        \centering
        \includegraphics[trim={0 0 2cm 0},clip,width=\textwidth]{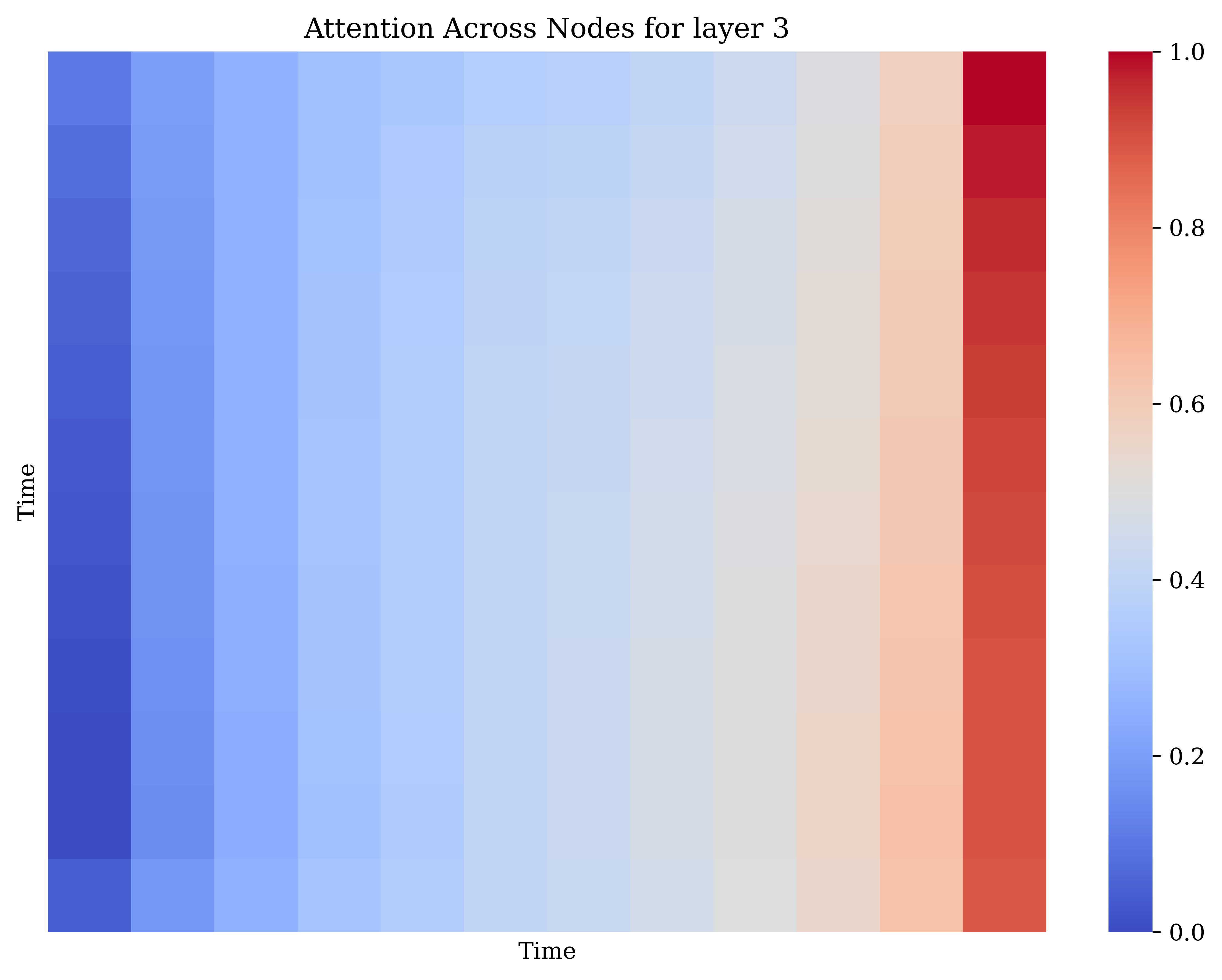}
        \caption{PEMS-BAY time-time attention scores layer 4}
    \end{subfigure}
    ~
    \begin{subfigure}[b]{0.45\textwidth}
        \centering
        \includegraphics[trim={0 0 2cm 0},clip,width=\textwidth]{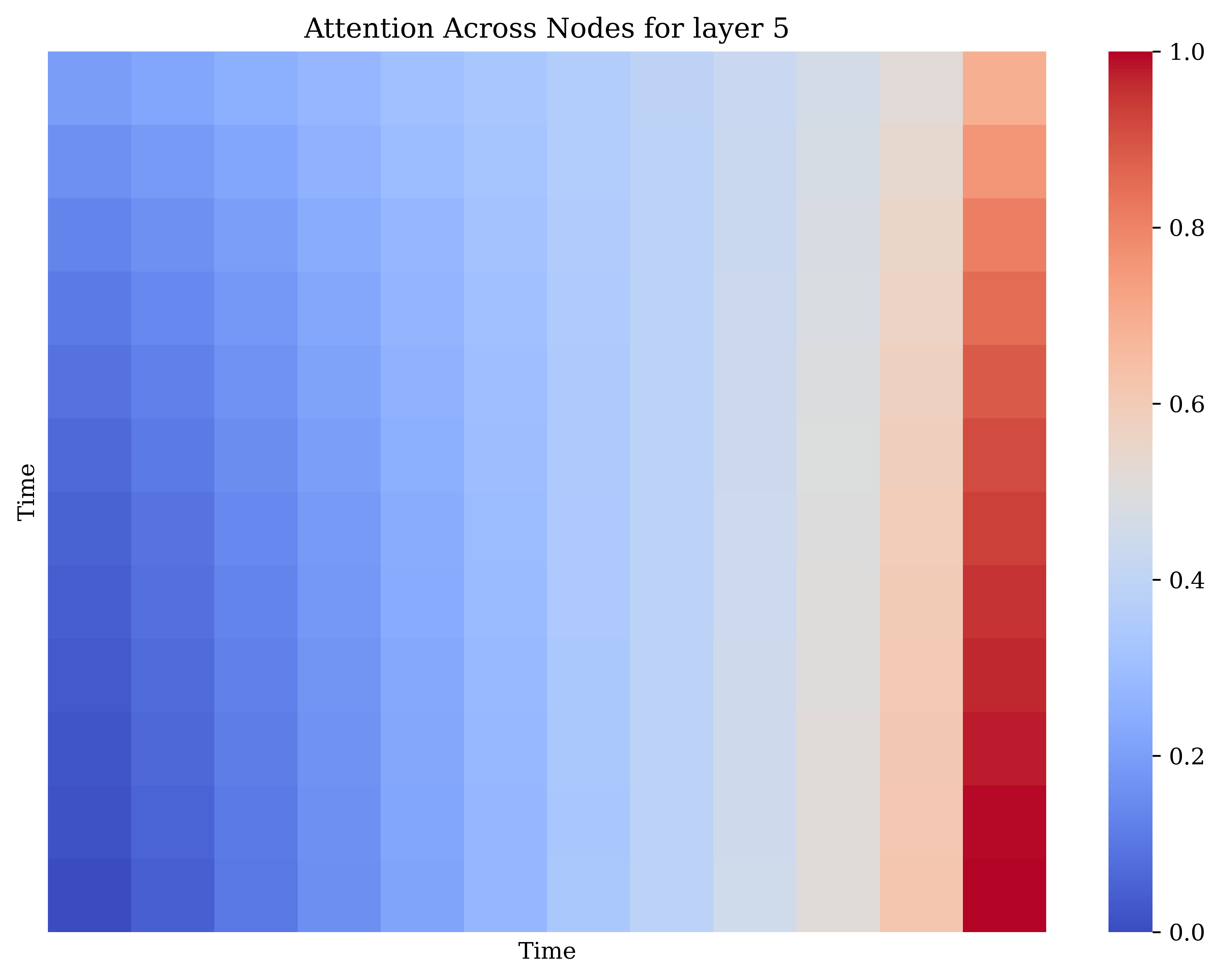}
        \caption{PEMS-BAY time-time attention scores layer 6}
    \end{subfigure}
    \caption{Attention heatmap reveals T-Graphormer learns spatial and temporal relationships simultaneously.}
    \label{fig:metr-attn}
\end{figure}
While the attention score heatmap averaged across layers shows how the model attends to nodes and time steps, the per-layer heatmap reveals interesting behaviour. For instance, the node-node attention scores across layers show the strongest self-attention. However, many nodes in the model's last layer actually attend to one node stronger (the bright red column on the left side) than their own values from previous timesteps. This is not selectivity is not present in the first layer. In the PEMS-BAY dataset, time steps in the fourth layer have increasingly stronger attention scores toward the right, whereas time steps in the last layer show a diagonal trend toward the top-right corner (except for the last column). Notice that the trend in the last column is also reversed. This shows that the model learns unique spatiotemporal patterns at each layer.

\subsubsection{Model Variant Performance}
\begin{table}[hbt]
    \centering
    \begin{tabular}{ccccc} \toprule
        \multirow{2}{*}{Dataset} & \multirow{2}{*}{Model} & \multicolumn{3}{c}{Horizon 12} \\ \cmidrule(lr){3-5} 
        & & MAE & RMSE & MAPE \\ \midrule
        
        \multirow{3}{*}{PEMS-BAY} 
        & micro   & 1.70  & 3.27  & 3.80  \\
        & \textbf{mini}    & \textbf{1.63}  & \textbf{3.20}  & \textbf{3.65}  \\
        & small   & 1.70  & 3.42  & 3.96  \\
        \midrule\midrule
        
        \multirow{3}{*}{METR-LA}  
        & micro             & 3.39  & 6.41  & 8.61  \\
        & \textbf{mini}           & \textbf{3.19}  & \textbf{6.12}  & \textbf{8.62} \\
        & small  &  3.39 & 6.36 & 8.78 \\\bottomrule
    \end{tabular}
    \caption{Additional results of T-Graphormer variants. The row with the best metric for each prediction length is bolded. The overall best-performing model for each dataset is also bolded.}
    \label{tab:add_pred_results}
\end{table}

\begin{table}[htb]
    \centering
    \resizebox{0.8\linewidth}{!}{%
    \begin{tabular}{ccccccc} \toprule
        \multirow{2}{*}{\textbf{Configuration}} & \multicolumn{3}{c}{\textbf{PEMS-BAY}} & \multicolumn{3}{c}{\textbf{METR-LA}} \\ \cmidrule(lr){2-4} \cmidrule(lr){5-7}
        & micro & mini & small & micro & mini & small \\ \midrule
        optimizer & \multicolumn{6}{c}{AdamW \citep{loshchilov2018decoupled}} \\
        optimizer momentum & \multicolumn{6}{c}{$\beta_1,\beta_2=0.9, 0.999$} \\
        learning rate schedule & \multicolumn{6}{c}{cosine decay \citep{loshchilov2017sgdr}} \\
        hidden dimension ($d$) & 64 & 128 & 192 & 64 & 128 & 192 \\
        epochs & 50 & 50 & 50 & 100 & 100 & 100 \\
        warmup epochs & 10 & 10 & 10 & 30 & 30 & 30 \\
        learning rate & 2.0e-3 & 1.0e-3 & 1.0e-3 & 2.0e-3 & 3.0e-3 & 2.0e-3 \\
        gradient clipping & 1.0 & 1.0 & 1.0 & 2.0 & 2.0 & 2.0 \\
        weigh decay & 1e-4 & 1e-4 & 1e-4 & 1e-4 & 1e-4 & 1e-4 \\
        batch size & 128 & 128 & 128 & 128 & 128 & 128 \\
        dropout & 0.1 & 0.1 & 0.1 & 0.1 & 0.1 & 0.1 \\
        layer-wise decay & 0.90 & 0.90 & 0.90 & 0.90 & 0.90 & 0.90 \\
        \# of parameters (M) & 0.58 & 1.76 & 4.44 & 0.49 & 1.57 & 4.15 \\\bottomrule
    \end{tabular}}
    \caption{Best training hyperparameters of T-Graphormer variants on PEMS-BAY and METR-LA.}
    \label{tab:training_hyperparameters}
\end{table}
We analyze the scalability of T-Graphormer by evaluating its size variants on the testing dataset. The results for the best-performing models are shown in Table ~\ref{tab:add_pred_results}. Its settings are detailed in Table~\ref{tab:training_hyperparameters}. We confirm that on both datasets, the mini model performs the best, and the small model performs the worst. Evidently, this is caused by overfitting, as shown in Figure~\ref{fig:train_val_diff_plot}. 

\begin{figure}
    \centering
    \begin{subfigure}[b]{0.45\textwidth}
        \centering
        \begin{tikzpicture}
            \definecolor{red}{RGB}{144,238,144}     
            \definecolor{green}{RGB}{100,144,100}     
            \definecolor{blue}{RGB}{0,128,0}     
            \definecolor{yellow}{RGB}{255,128,0}   
            \definecolor{orange}{RGB}{255,128,100}   
            \definecolor{purple}{RGB}{200, 162, 200} 

            \begin{axis}[
                title={},
                width=\textwidth,
                xlabel={Epochs},
                ylabel={Metric},
                xmin=0, xmax=50,
                ymin=1, ymax=11,  
                xtick={0,10,20,30,40,50},
                ytick={1,2,4,6,8,10},
                legend pos=north east,
                legend style={
                    nodes={scale=0.5, transform shape},
                    line width=.3pt,
                    mark size=.4pt
                }, 
                ymajorgrids=true,
                grid style=dashed,
            ]
            \addplot[
                color=orange,
                mark=o,
                mark size=1pt
                ]
                coordinates {(1,3.5766190240093825)(2,3.5447350910731723)(3,3.459993422260299)(4,3.3685906963872103)(5,3.1924062417491057)(6,3.0314186387988644)(7,2.899035053639551)(8,2.7870259625113323)(9,2.669611443206096)(10,2.4478928055356723)(11,2.506105776481365)(12,2.25983641792186)(13,2.286104467515755)(14,2.2680610424271013)(15,2.1295260378658862)(16,1.9569207914230824)(17,1.9508840906821452)(18,1.807606479401962)(19,1.7966644482403855)(20,1.7847829120499747)(21,1.7798973428580436)(22,1.772076274621688)(23,1.7790907157639388)(24,1.7588939299140292)(25,1.719504092008837)(26,1.677700085070459)(27,1.6942909944075777)(28,1.660139100709086)(29,1.670241833786078)(30,1.6507929341034957)(31,1.65852987171135)(32,1.662157164420217)(33,1.65280284580364)(34,1.653417579955586)(35,1.6297782370663276)(36,1.6452013361746998)(37,1.6390384455338785)(38,1.6516842406740937)(39,1.630897664620946)(40,1.6324010141159533)(41,1.6490381797246303)(42,1.6235298523064217)(43,1.623972482624508)(44,1.6355777317386253)(45,1.622101947367649)(46,1.6328784822044284)(47,1.6215183224454637)(48,1.6185275394063208)(49,1.6269573192076383)(50,1.6278034506305572)
};
            
            \addplot[
                color=green,
                mark=square,
                mark size=1pt
                ]
                coordinates {(1,6.002044227449209)(2,5.943720592057101)(3,6.040633838389144)(4,5.838495302951098)(5,5.456026325852091)(6,5.400754887295942)(7,5.159077937610322)(8,4.805065736975721)(9,4.84175718737088)(10,4.442661379156391)(11,4.398360503983388)(12,4.152358040007578)(13,4.167239620388927)(14,4.216690243022966)(15,3.901724860964832)(16,3.6697982319855287)(17,3.5524053050808826)(18,3.4627198846109453)(19,3.485757363282041)(20,3.3852381076314666)(21,3.4230047494585065)(22,3.384937602711896)(23,3.402042885987623)(24,3.3379033402913176)(25,3.37058671581031)(26,3.3024429625812948)(27,3.3187659586995792)(28,3.280658261109424)(29,3.2772235094860034)(30,3.2814605015397253)(31,3.25816418141264)(32,3.262524632905852)(33,3.2733022268466687)(34,3.267637986832866)(35,3.2434698474205765)(36,3.276887117168321)(37,3.2623167311724064)(38,3.270344426829694)(39,3.2414019940086223)(40,3.2298577863682985)(41,3.2646837292178983)(42,3.23497461558487)(43,3.2388110929004243)(44,3.243674877052483)(45,3.2342555977964915)(46,3.247761874704317)(47,3.235918727155472)(48,3.2307853282138868)(49,3.243004058546368)(50,3.235937119628977)
};

            \addplot[
                color=purple,
                mark=diamond,
                mark size=1pt
                ]
                coordinates {(1,10.51571928851661)(2,10.2745914939768)(3,9.28187607136625)(4,8.95950820435771)(5,9.023949301969001)(6,8.101641613426409)(7,7.862657825884541)(8,7.62775256441542)(9,7.49596831776465)(10,6.2348249329743295)(11,6.274117608146939)(12,5.52200892526719)(13,5.55970735115481)(14,5.92795894109284)(15,5.15520177765845)(16,4.8780939815002196)(17,4.5713744237132)(18,4.19825546195968)(19,4.14933192725372)(20,4.21324639277856)(21,4.1082667582417995)(22,4.31136163237709)(23,4.28343446476073)(24,4.10163819842818)(25,3.9658198654565804)(26,3.79291465162612)(27,3.75438218395556)(28,3.7321072649038998)(29,3.81514448550931)(30,3.81079969095057)(31,3.77442214892642)(32,3.75051120476376)(33,3.6652304633404604)(34,3.75321113494455)(35,3.71235275652528)(36,3.7249690147771997)(37,3.69197862156982)(38,3.80094552622927)(39,3.6781093862462697)(40,3.70847853043684)(41,3.7817400429517902)(42,3.64983829271946)(43,3.68467515374543)(44,3.6586911101380597)(45,3.6311336203596998)(46,3.64723743222791)(47,3.6354670808228495)(48,3.6582566081716403)(49,3.66636854975545)(50,3.67521833665611)
};
            \legend{MAE, RMSE, MAPE (\%)}
                
            \end{axis}
        \end{tikzpicture}
        \caption{PEMS-BAY}
    \end{subfigure}
    ~
    \begin{subfigure}[b]{0.45\textwidth}
        \centering
        \begin{tikzpicture}
            \definecolor{red}{RGB}{144,238,144}     
            \definecolor{green}{RGB}{100,144,100}     
            \definecolor{blue}{RGB}{0,128,0}     
            \definecolor{yellow}{RGB}{255,128,0}   
            \definecolor{orange}{RGB}{255,128,100}   
            \definecolor{purple}{RGB}{200, 162, 200}

            \begin{axis}[
                title={},
                width=\textwidth,
                xlabel={Epochs},
                ylabel={Metric},
                xmin=0, xmax=100,
                ymin=2, ymax=16,  
                xtick={0,25,50,75,100},
                ytick={2,3,6,9,12,15},
                legend pos=north east,
                legend style={
                    nodes={scale=0.5, transform shape},
                    line width=.3pt,
                    mark size=.4pt
                },  
                ymajorgrids=true,
                grid style=dashed,
            ]
            
            \addplot[
                color=orange,
                mark=o,
                mark size=1pt
                ]
                coordinates {(1,4.910949649341474)(2,4.9799577470599)(3,5.057421638029758)(4,4.96293240642854)(5,4.9359968667675815)(6,4.887386749520306)(7,4.684018579868722)(8,4.776752442256417)(9,4.275731681924418)(10,4.047931984114835)(11,3.3453846508245397)(12,3.7016547673825766)(13,3.809163871025405)(14,3.2771565646899288)(15,3.284714805741281)(16,3.289963214488474)(17,3.378773108673082)(18,3.1823588275420263)(19,3.181584701002967)(20,3.098597352954376)(21,3.194443126966324)(22,3.251144160181458)(23,3.327084396243374)(24,3.031056641727233)(25,3.1858042032484)(26,3.146529861547436)(27,3.181063439757955)(28,3.0971936130304343)(29,3.187783450619361)(30,3.290889295805942)(31,3.1889337434294136)(32,3.2684441693127155)(33,3.1121739036203406)(34,3.050864168795296)(35,3.0287683635349514)(36,3.1887288100225417)(37,3.101710790154637)(38,3.131181866800952)(39,3.0288145426991524)(40,3.0068159755493435)(41,3.052690072838113)(42,3.133973128749841)(43,3.055966158175545)(44,3.0895266811728965)(45,3.0158126729227184)(46,3.11267391411222)(47,2.9466826096742857)(48,3.0464478798045245)(49,3.0945642877666053)(50,3.1583007860310794)(51,3.037362026163838)(52,3.1228425419610866)(53,3.068862908293502)(54,3.173287904603737)(55,3.122241956939068)(56,3.064923480270587)(57,3.1135312163175266)(58,3.031352247829575)(59,3.1038517800037897)(60,3.0409612627787013)(61,3.0034487643192094)(62,3.016346446077435)(63,3.0367939710860776)(64,3.0016139204413674)(65,3.0730739203629858)(66,3.025208413674905)(67,3.0422880660751224)(68,3.0202369718969075)(69,3.0565480796019675)(70,3.065300175075846)(71,2.9954734720446377)(72,3.022574786331818)(73,3.0431022605005755)(74,3.0165664314762384)(75,3.001943832694196)(76,2.9892928016755)(77,3.0210829833000203)(78,3.005838851331272)(79,3.0072709012115117)(80,3.006626837059995)(81,3.0081532743942234)(82,3.0268730906395365)(83,3.023882388285248)(84,3.008884074510223)(85,3.0084846294485037)(86,3.020354284487108)(87,3.0006270784989137)(88,3.019334751755765)(89,3.003915723870186)(90,3.013118324222513)(91,2.9921401631322078)(92,3.010637593425149)(93,3.004394894504972)(94,3.001538177203631)(95,3.006154445113144)(96,3.007028010343071)(97,2.999526775851149)(98,2.9955652613828185)(99,2.997945813657977)(100,3.0060843074276486)
};
            
            \addplot[
                color=green,
                mark=square,
                mark size=1pt
                ]
                coordinates {(1,8.437626071552806)(2,8.495477173930016)(3,8.643447852998136)(4,8.592226626717041)(5,8.52894025567536)(6,8.820661554687492)(7,8.286901454362914)(8,7.9326633531078)(9,7.690748604031924)(10,7.164609144572343)(11,6.08574460335424)(12,6.789697822148555)(13,7.114875967640464)(14,6.0041409835706805)(15,5.991794899001579)(16,6.0379649671229805)(17,6.107657220284237)(18,5.891882323146423)(19,5.920920889824629)(20,5.681167065749102)(21,5.867538765351349)(22,5.970770695011749)(23,5.92325623873099)(24,5.705223210295227)(25,5.953769900252886)(26,5.838855159790995)(27,5.8331650559242085)(28,5.82815898418705)(29,5.816260265064574)(30,6.194288253993075)(31,5.861975766718388)(32,5.998989245253745)(33,5.847180966949352)(34,5.697392199292918)(35,5.805245410616153)(36,6.034954161148205)(37,5.925689154487347)(38,5.843869179347966)(39,5.775133729329176)(40,5.782193048186948)(41,5.86567624693162)(42,5.967315169730198)(43,5.939517558804739)(44,6.0007010996341705)(45,5.769734030293527)(46,5.985261243637477)(47,5.642294469524488)(48,5.705022061261062)(49,5.945689106133775)(50,6.041926555857759)(51,5.853658847789341)(52,6.050800099551121)(53,5.858206844636213)(54,6.076322696685234)(55,6.015817091008213)(56,5.903817927934856)(57,5.9795451987987365)(58,5.875142674858325)(59,5.989945496876385)(60,5.905345140355769)(61,5.775891830883572)(62,5.823465961928123)(63,5.845337604807916)(64,5.745089691445649)(65,5.906380210106618)(66,5.853694292150926)(67,5.915228257739098)(68,5.830286784950539)(69,5.91209153566405)(70,5.891985266217005)(71,5.765318132797692)(72,5.843841390562392)(73,5.903647854158255)(74,5.799336903215012)(75,5.790133822535243)(76,5.799566809540597)(77,5.839836824163098)(78,5.819061985813847)(79,5.798546042850363)(80,5.809297324584745)(81,5.821500429901008)(82,5.841493836664986)(83,5.84943446266317)(84,5.836522778095764)(85,5.825310739714687)(86,5.860976592019618)(87,5.819108583799032)(88,5.858603566049416)(89,5.8182119325012245)(90,5.856450992000994)(91,5.7909700655352285)(92,5.840301260978819)(93,5.822644235897008)(94,5.826053487154368)(95,5.835999826021562)(96,5.831592925275876)(97,5.815802389285832)(98,5.812289919564936)(99,5.811047585977015)(100,5.832435145660817)
};

            \addplot[
                color=purple,
                mark=diamond,
                mark size=1pt
                ]
                coordinates {(1,14.27110550720048)(2,14.84792658926703)(3,14.894745758599731)(4,14.88606035611461)(5,13.92857266872061)(6,13.16689810780623)(7,13.26061258931365)(8,13.54089977526243)(9,12.52986330050207)(10,10.97099388470583)(11,9.41221442475597)(12,9.11993117061624)(13,9.200820254766331)(14,8.59974726496556)(15,8.35278874981787)(16,8.20547167627112)(17,8.44785182155553)(18,8.25257057525466)(19,8.34947910555448)(20,7.9319808472676305)(21,8.13761766325131)(22,8.51370950411569)(23,8.30814923962445)(24,7.99644780675059)(25,8.209054279288)(26,7.99629467018068)(27,8.12363405573909)(28,8.0569823138567)(29,8.23041482620614)(30,8.29681027477996)(31,8.44114874839803)(32,8.30225374595323)(33,8.47348926663556)(34,8.06818962931251)(35,8.12150602935183)(36,8.27405655125012)(37,8.39094089460559)(38,8.037844720596029)(39,8.3132301846388)(40,7.94115998879728)(41,8.10819222555369)(42,8.03973487308782)(43,8.171384978219889)(44,8.19500185229389)(45,7.836975213979359)(46,8.07856920752027)(47,8.07442537273082)(48,8.40855510999904)(49,8.08092771683086)(50,8.17176916903901)(51,8.19254280468498)(52,8.014158983227999)(53,8.31783814186412)(54,8.22291145318516)(55,8.30695834256998)(56,8.20385669143195)(57,8.33404620974863)(58,8.11177914716421)(59,8.06582769078869)(60,8.4357671196617)(61,8.255238366068749)(62,8.11085864546176)(63,8.17427594355029)(64,8.19373355957948)(65,8.15905780004741)(66,8.20364325494852)(67,8.24765772720587)(68,8.25227888124351)(69,8.152687451515229)(70,8.17620967174543)(71,8.094397819112489)(72,8.21112914555695)(73,8.14823425180718)(74,8.16132294820027)(75,8.13320350684862)(76,8.18773978693757)(77,8.24224282731057)(78,8.21721824140355)(79,8.222288756346261)(80,8.161952651670761)(81,8.24231693954843)(82,8.21852697921475)(83,8.230130316754341)(84,8.223952057759961)(85,8.24843612498855)(86,8.22793891564747)(87,8.21268879203289)(88,8.198475596744121)(89,8.231830779767959)(90,8.25508646163806)(91,8.20387064693626)(92,8.21018417582903)(93,8.2542664518786)(94,8.23090883214913)(95,8.22639903991355)(96,8.23968310118136)(97,8.23344072977366)(98,8.23503261879217)(99,8.214729347830101)(100,8.24111864536105)
};
            \legend{MAE, RMSE, MAPE (\%)}
            \end{axis}
        \end{tikzpicture}
        \caption{METR-LA}
    \end{subfigure}
    \caption{Validation metrics on two traffic prediction datasets. The best-performing model is displayed.}
    \label{fig:val_plot}
\end{figure}
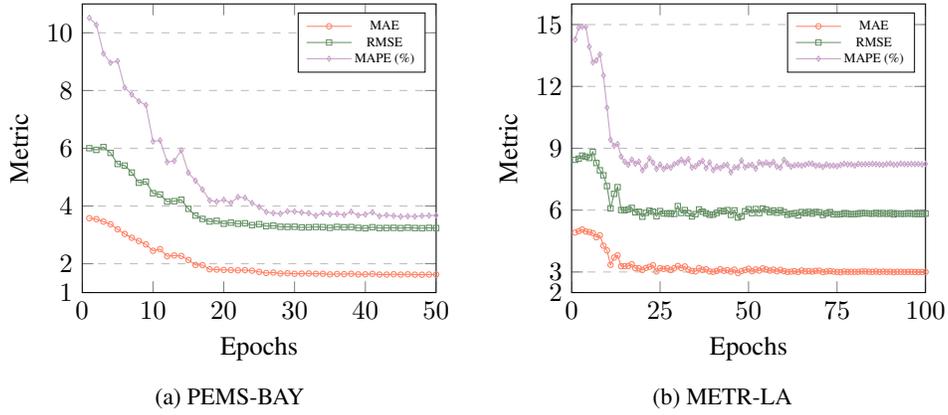
\newcommand\boxsize{0.85}

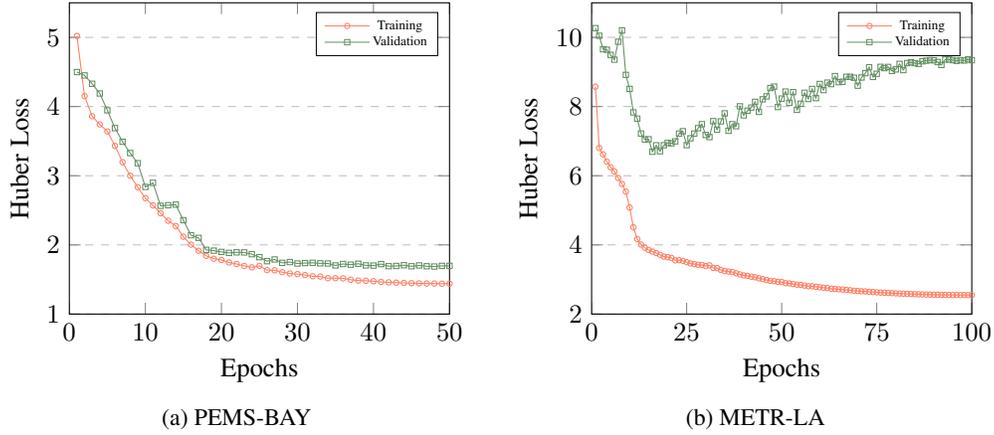
\begin{figure}
    \centering
    \begin{subfigure}[b]{0.475\textwidth}
        \centering
        \begin{tikzpicture}
            \definecolor{red}{RGB}{144,238,144}     
            \definecolor{green}{RGB}{100,144,100}     
            \definecolor{blue}{RGB}{0,128,0}     
            \definecolor{yellow}{RGB}{255,128,0}   
            \definecolor{orange}{RGB}{255,128,100}   
            \begin{axis}[
                title={},
                width=\textwidth,
                xlabel={Epochs},
                ylabel={Huber Loss},
                xmin=0, xmax=50,
                ymin=1, ymax=5.5,  
                xtick={0,10,20,30,40,50},
                ytick={1,2,3,4,5},
                legend pos=north east,
                legend style={
                    nodes={scale=0.5, transform shape},
                    line width=.3pt,
                    mark size=.4pt
                }, 
                ymajorgrids=true,
                grid style=dashed,
            ]
            \addplot[
                color=orange,
                mark=o,
                mark size=1pt
                ]
                coordinates {(1,5.020945832892088)(2,4.150326182069854)(3,3.8595888955458095)(4,3.742192481586397)(5,3.640826785127461)(6,3.432603089634372)(7,3.1958417429133297)(8,2.999660747611439)(9,2.834118825148814)(10,2.675866559239413)(11,2.5705909936167988)(12,2.4577006201101983)(13,2.349578441072442)(14,2.2733583634717616)(15,2.117806035897729)(16,2.003602786701012)(17,1.9154034670707267)(18,1.8415369986506596)(19,1.800153142559204)(20,1.782809933439826)(21,1.7460389472838551)(22,1.7238230574761595)(23,1.6962180475069044)(24,1.6761644210329538)(25,1.696676407444362)(26,1.6354609752716571)(27,1.6329154540883597)(28,1.6057488284678023)(29,1.5866204459478557)(30,1.5779495934699843)(31,1.5640927278669756)(32,1.5469326805863022)(33,1.5424519741451457)(34,1.5189609529517585)(35,1.5190749073513932)(36,1.5169858521404918)(37,1.496450931370206)(38,1.4847210680639096)(39,1.482044617356806)(40,1.4750058488320164)(41,1.4649903416090675)(42,1.4586644351116076)(43,1.454338256322024)(44,1.4505568815851495)(45,1.4473973225877408)(46,1.444537411763597)(47,1.4424637699854594)(48,1.4416539406696625)(49,1.439547751347442)(50,1.4404156008069715)
};
            
            \addplot[
                color=green,
                mark=square,
                mark size=1pt
                ]
                coordinates {(1,4.498456223723343)(2,4.449846162774047)(3,4.329440921865484)(4,4.189183503755593)(5,3.94598170107777)(6,3.688635485695987)(7,3.491127283342423)(8,3.3279962025113554)(9,3.181668630530757)(10,2.8382214619770942)(11,2.897211896292808)(12,2.565413712951628)(13,2.573999608716657)(14,2.582219851593817)(15,2.357282857145948)(16,2.1403479580917666)(17,2.101793418053292)(18,1.9281825482982644)(19,1.91718848940048)(20,1.8985165636843435)(21,1.8841219774680569)(22,1.8925314955585013)(23,1.891163641727099)(24,1.8660234957475632)(25,1.8228333237396408)(26,1.7648524395011354)(27,1.7894086363037245)(28,1.7404109777713883)(29,1.752065245800304)(30,1.7308862182996974)(31,1.7375789338954584)(32,1.7409551896257884)(33,1.733474342092391)(34,1.7314568819355125)(35,1.7031208983581003)(36,1.7239209798303434)(37,1.7145967837852267)(38,1.7276509497801096)(39,1.7035672662124473)(40,1.7011006400332471)(41,1.722889287474518)(42,1.6922702275434032)(43,1.695101808521971)(44,1.706282476034765)(45,1.6912168945813877)(46,1.7032733154003887)(47,1.6910227310776529)(48,1.6873227802839148)(49,1.6966864882114296)(50,1.6966554557535505)
};
            
            \legend{Training, Validation}
                
            \end{axis}
        \end{tikzpicture}
        \caption{PEMS-BAY}
    \end{subfigure}
    ~
    \begin{subfigure}[b]{0.475\textwidth}
        \centering
        \begin{tikzpicture}
            \definecolor{red}{RGB}{144,238,144}     
            \definecolor{green}{RGB}{100,144,100}     
            \definecolor{blue}{RGB}{0,128,0}     
            \definecolor{yellow}{RGB}{255,128,0}   
            \definecolor{orange}{RGB}{255,128,100}   
            
            \begin{axis}[
                title={},
                width=\textwidth,
                xlabel={Epochs},
                ylabel={Huber Loss},
                xmin=0, xmax=100,
                ymin=2, ymax=11,  
                xtick={0,25,50,75,100},
                ytick={2,4,6,8,10},
                legend pos=north east,
                legend style={
                    nodes={scale=0.5, transform shape},
                    line width=.3pt,
                    mark size=.4pt
                },  
                ymajorgrids=true,
                grid style=dashed,
            ]
            
            \addplot[
                color=orange,
                mark=o,
                mark size=1pt
                ]
                coordinates {(1,8.574798453952345)(2,6.800290892853119)(3,6.617943632706145)(4,6.405624701518401)(5,6.2443935863166375)(6,6.124076225609582)(7,5.933289807613129)(8,5.762525934163973)(9,5.541826949896259)(10,5.080785126727477)(11,4.508461785133436)(12,4.16776226315702)(13,4.009976981120689)(14,3.9174316616179303)(15,3.853822216570616)(16,3.809173044578256)(17,3.7636565085405977)(18,3.7110691012265047)(19,3.661032719768892)(20,3.645793878009546)(21,3.6248421370784176)(22,3.5546280680415787)(23,3.557478626197744)(24,3.53956422185866)(25,3.502994596659739)(26,3.4641669225668874)(27,3.444480610665715)(28,3.428025451437813)(29,3.4199706874240703)(30,3.391105899186892)(31,3.409211529311892)(32,3.336347538336256)(33,3.329471149376142)(34,3.2751911728579466)(35,3.243901619744874)(36,3.229833461234821)(37,3.213817268331474)(38,3.179317842159316)(39,3.147059843918988)(40,3.117035429332858)(41,3.1077487486425803)(42,3.079285559949474)(43,3.0651656921818677)(44,3.0356623201527806)(45,3.012108727613183)(46,2.9842997789183987)(47,2.9624649911959757)(48,2.951520600210681)(49,2.932841424689751)(50,2.9275324962167777)(51,2.893568542436541)(52,2.888872773688689)(53,2.8675092634196595)(54,2.851613438535278)(55,2.845055787820364)(56,2.8250887611400937)(57,2.806797016227993)(58,2.805313214719375)(59,2.7891086568740087)(60,2.7768629410556542)(61,2.763877404309083)(62,2.7540054121744806)(63,2.7340442314445577)(64,2.7281200481710512)(65,2.716132979427224)(66,2.710443916005668)(67,2.7009696296561705)(68,2.689914904166764)(69,2.675088712445725)(70,2.668659556831155)(71,2.6593579383534647)(72,2.654489794325765)(73,2.643415582713202)(74,2.6376566292844883)(75,2.630012423416801)(76,2.622842138376987)(77,2.618559128173362)(78,2.610458186356343)(79,2.6061394184788016)(80,2.600523534102816)(81,2.5937225669224526)(82,2.5887459401732613)(83,2.5852684247915514)(84,2.580566316365559)(85,2.5765948186371133)(86,2.5727264066046485)(87,2.5686049337619457)(88,2.56515808286113)(89,2.563118528559784)(90,2.5605293854732856)(91,2.5579447078808286)(92,2.5558243374799057)(93,2.554750759189374)(94,2.553045452934241)(95,2.5520957444753445)(96,2.5510766352209773)(97,2.549633083061796)(98,2.549640311398239)(99,2.549275276418999)(100,2.54879930740762)
};
            
            \addplot[
                color=green,
                mark=square,
                mark size=1pt
                ]
                coordinates {(1,10.267160965460484)(2,10.047169962636778)(3,9.65730015905661)(4,9.643011372501604)(5,9.49350481041681)(6,9.3558375490046)(7,9.87433505295036)(8,10.19955750937774)(9,8.916200666366336)(10,8.509399003236092)(11,7.830533229441286)(12,7.65097001459554)(13,7.220081196349358)(14,7.048375919043461)(15,7.053779386610628)(16,6.692760035117096)(17,6.879866760746341)(18,6.707145220695812)(19,6.882234955223922)(20,6.949331183996156)(21,6.936427037988867)(22,6.993526822495683)(23,7.220268892768387)(24,7.286530090130378)(25,6.884135139183464)(26,7.082284753289178)(27,7.219333121431208)(28,7.374787622801612)(29,7.489525484286736)(30,7.178299986034911)(31,7.117836260349951)(32,7.577585284258718)(33,7.332336522917324)(34,7.571380932824077)(35,7.803166471909139)(36,7.299819571671085)(37,7.490439226903091)(38,7.431533591630303)(39,8.006853898830503)(40,7.746234128129816)(41,7.88044035003007)(42,7.970853225162653)(43,8.136044321355419)(44,7.846471107939136)(45,8.209490200696148)(46,8.29423891175016)(47,8.535770606145123)(48,8.576225228120233)(49,7.987256987699282)(50,8.229673485193297)(51,8.43311682127625)(52,8.109916548444845)(53,8.416842596135407)(54,7.90497683176649)(55,8.080502321229917)(56,8.398101468091813)(57,8.218785006169961)(58,8.508100563956198)(59,8.243315694975518)(60,8.652768132564063)(61,8.47593410730919)(62,8.692092642536231)(63,8.649674479510182)(64,8.87917689477729)(65,8.705012497083048)(66,8.716145039718842)(67,8.85818762548059)(68,8.859931251364891)(69,8.833311328890725)(70,8.603550182255072)(71,8.839631576891815)(72,8.965721004755698)(73,9.136386454871325)(74,8.85625428065797)(75,8.947260942136015)(76,9.14791689938474)(77,9.110563209680754)(78,9.144739753592797)(79,9.029382900796204)(80,9.073433872904175)(81,9.233273554071088)(82,9.055787253741906)(83,9.256543570520163)(84,9.279100959824625)(85,9.255935024087115)(86,9.233545532084516)(87,9.307979107251237)(88,9.32688337940479)(89,9.34193211341294)(90,9.33792324392038)(91,9.287487206963178)(92,9.202479559893362)(93,9.359561769413611)(94,9.360029674718312)(95,9.34230448222049)(96,9.321430067731956)(97,9.341384628944308)(98,9.331146094773974)(99,9.364312022088845)(100,9.340156876664851)
};
            \legend{Training, Validation}
            \end{axis}
        \end{tikzpicture}
        \caption{METR-LA}
    \end{subfigure}
    \caption{Training and validation loss of the best-performing models on two traffic prediction datasets. Training loss is plotted as the global average per epoch, and validation loss is computed as the average per epoch. Both datasets use Huber loss with $\delta=1.5$. For PEMS-BAY, the gap between training and validation is much closer than that of METR-LA. This is further shown in Figure ~\ref{fig:train_val_diff_plot}.}
    \label{fig:loss_plot}
\end{figure}
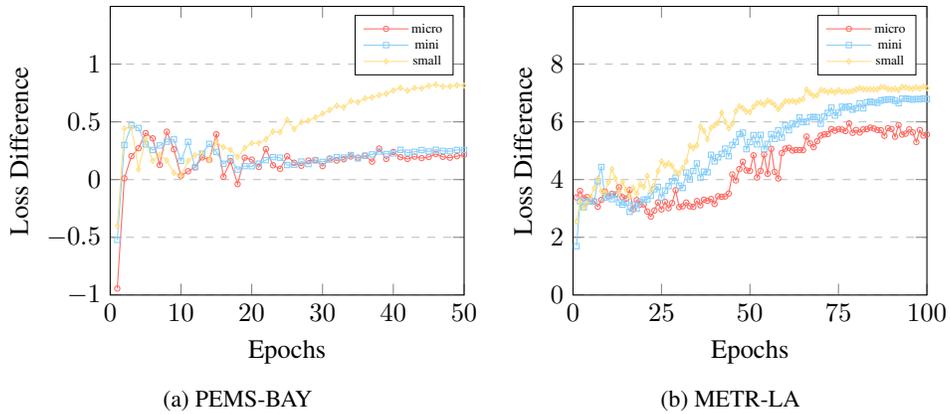
\begin{figure}
    \centering
    \begin{subfigure}[b]{0.45\textwidth}
        \centering
        \begin{tikzpicture}
            \definecolor{red}{RGB}{255,105,97}      
            \definecolor{green}{RGB}{119,221,119}   
            \definecolor{blue}{RGB}{135,206,250}    
            \definecolor{yellow}{RGB}{255,223,128}  
            \definecolor{orange}{RGB}{255,179,102}  
            \definecolor{purple}{RGB}{216,191,216}  

            \begin{axis}[
                title={},
                width=\textwidth,
                xlabel={Epochs},
                ylabel={Loss Difference},
                xmin=0, xmax=50,
                ymin=-1, ymax=1.5,  
                xtick={0,10,20,30,40,50},
                ytick={-1,-0.5,0,0.5,1},
                legend pos=north east,
                legend style={
                    nodes={scale=0.5, transform shape},
                    line width=.3pt,
                    mark size=.4pt
                }, 
                ymajorgrids=true,
                grid style=dashed,
            ]
            \addplot[
                color=red,
                mark=o,
                mark size=1pt
                ]
                coordinates {(1,-0.9432438535356917)(2,0.01081080192690731)(3,0.20333540092198277)(4,0.2711715505892265)(5,0.4030311596344389)(6,0.3568185891238733)(7,0.1267775855010087)(8,0.41481983022965396)(9,0.2620959415099433)(10,0.03310492108441254)(11,0.07203098593544821)(12,0.10638248886887025)(13,0.19239925550229886)(14,0.17065405496707298)(15,0.3922516257184432)(16,0.02480184165328514)(17,0.15925277941555072)(18,-0.03883298997669127)(19,0.1896385217786809)(20,0.17273628283817666)(21,0.10842112426520134)(22,0.2641313357969868)(23,0.12403549771378364)(24,0.09337333365799672)(25,0.20200910072491807)(26,0.1460251649363029)(27,0.12013982870056172)(28,0.16480554085628474)(29,0.16990639514610129)(30,0.1164470329403633)(31,0.18254609125845112)(32,0.16513965579201928)(33,0.1759346816140812)(34,0.20047409399202687)(35,0.19033730304743202)(36,0.20707992104837736)(37,0.15554903330176817)(38,0.2685678964262066)(39,0.17879666140182615)(40,0.23934480242706968)(41,0.19388798379544037)(42,0.17919135548626586)(43,0.19928891896214118)(44,0.18581569032534428)(45,0.19566083863629014)(46,0.22223701706842536)(47,0.19709940740104437)(48,0.1895863333067156)(49,0.20301042260365643)(50,0.214992917230038)
};
            
            \addplot[
                color=blue,
                mark=square,
                mark size=1pt
                ]
                coordinates {(1,-0.522489609168745)(2,0.2995199807041926)(3,0.4698520263196748)(4,0.4469910221691955)(5,0.30515491595030886)(6,0.2560323960616149)(7,0.2952855404290933)(8,0.32833545489991645)(9,0.34754980538194324)(10,0.16235490273768116)(11,0.3266209026760092)(12,0.1077130928414296)(13,0.22442116764421494)(14,0.3088614881220555)(15,0.239476821248219)(16,0.13674517139075437)(17,0.18638995098256506)(18,0.08664554964760485)(19,0.11703534684127614)(20,0.11570663024451755)(21,0.13808303018420176)(22,0.16870843808234182)(23,0.19494559422019453)(24,0.1898590747146094)(25,0.1261569162952787)(26,0.12939146422947823)(27,0.1564931822153648)(28,0.134662149303586)(29,0.16544479985244842)(30,0.15293662482971304)(31,0.17348620602848275)(32,0.19402250903948626)(33,0.19102236794724536)(34,0.21249592898375402)(35,0.18404599100670715)(36,0.20693512768985167)(37,0.21814585241502082)(38,0.2429298817162)(39,0.2215226488556412)(40,0.22609479120123077)(41,0.2578989458654506)(42,0.23360579243179558)(43,0.24076355219994694)(44,0.2557255944496155)(45,0.2438195719936469)(46,0.2587359036367918)(47,0.24855896109219344)(48,0.24566883961425234)(49,0.2571387368639877)(50,0.25623985494657897)
};

            \addplot[
                color=yellow,
                mark=diamond,
                mark size=1pt
                ]
                coordinates {(1,-0.4014971454267906)(2,0.4419612159083117)(3,0.4535963381343251)(4,0.08841444069300142)(5,0.34832340952712837)(6,0.16226013280927098)(7,0.21454763416399514)(8,0.16628490683179376)(9,0.0580826125538727)(10,0.03464084099475517)(11,0.16335756952155145)(12,0.211084850443928)(13,0.22642453294186082)(14,0.17040684744666645)(15,0.32478077695205276)(16,0.27994832298654027)(17,0.25839481987132706)(18,0.1936833516175296)(19,0.27316506712932287)(20,0.31497944871833106)(21,0.3163615314681094)(22,0.34911919058772045)(23,0.4165060350878642)(24,0.4092675404962818)(25,0.5169786829513647)(26,0.4369951655390505)(27,0.4972078531951085)(28,0.5110247698051509)(29,0.5378692650639005)(30,0.5709121482884556)(31,0.6022571228699631)(32,0.6390287195693474)(33,0.6260181344000817)(34,0.6811219752668707)(35,0.6703013708131964)(36,0.7032857666097438)(37,0.7119242013104068)(38,0.7311413016432677)(39,0.7436584303691403)(40,0.7715936541665223)(41,0.7951111932965076)(42,0.7697383461616947)(43,0.7936537936893908)(44,0.7908667498479713)(45,0.8113327732087751)(46,0.8228081886272159)(47,0.805642077102129)(48,0.807627122886013)(49,0.8180179256698148)(50,0.8139161220098734)
};
            \legend{micro, mini, small}
                
            \end{axis}
        \end{tikzpicture}
        \caption{PEMS-BAY}
    \end{subfigure}
    ~
    \begin{subfigure}[b]{0.45\textwidth}
        \centering
        \begin{tikzpicture}
            \definecolor{red}{RGB}{255,105,97}      
            \definecolor{green}{RGB}{119,221,119}   
            \definecolor{blue}{RGB}{135,206,250}    
            \definecolor{yellow}{RGB}{255,223,128}  
            \definecolor{orange}{RGB}{255,179,102}  
            \definecolor{purple}{RGB}{216,191,216}  

            \begin{axis}[
                title={},
                width=\textwidth,
                xlabel={Epochs},
                ylabel={Loss Difference},
                xmin=0, xmax=100,
                ymin=0, ymax=10,  
                xtick={0,25,50,75,100},
                ytick={0,2,4,6,8},
                legend pos=north east,
                legend style={
                    nodes={scale=0.5, transform shape},
                    line width=.3pt,
                    mark size=.4pt
                },  
                ymajorgrids=true,
                grid style=dashed,
            ]
            
            \addplot[
                color=red,
                mark=o,
                mark size=1pt
                ]
                coordinates {(1,3.3819160828556925)(2,3.6096413112285344)(3,3.351567635263545)(4,3.3942665421935727)(5,3.243577980286573)(6,3.247844549688982)(7,3.057453088805379)(8,3.295138398214031)(9,3.5835863026870776)(10,3.425532631203585)(11,3.505595799549269)(12,3.482804076334193)(13,3.7315736169965428)(14,3.3858261053485137)(15,3.34033198574258)(16,3.6382259730411652)(17,2.9704893023063637)(18,3.2902883973367354)(19,3.13581611175906)(20,3.1384470917637213)(21,2.8932517490943557)(22,2.7152297158426366)(23,2.9284444090467705)(24,3.205951194157252)(25,2.9652624302164288)(26,3.2266252167261515)(27,3.0100434188340124)(28,3.1867584192692773)(29,3.632966782829064)(30,3.0468500239689043)(31,3.073622024671917)(32,3.2034079281852534)(33,3.0761427803520114)(34,3.0551286341104262)(35,3.257753121947261)(36,3.1090418063543495)(37,3.3044021317505177)(38,3.264347572062706)(39,3.326075473926433)(40,3.1538497256599745)(41,3.4223692772685177)(42,3.405031996011232)(43,3.41557906287292)(44,3.515608658902138)(45,4.1789227450702935)(46,3.9716866836489464)(47,4.363281520492637)(48,4.615903340099436)(49,4.31787869448977)(50,4.30126595402233)(51,4.841088228642954)(52,4.072579775681907)(53,4.303435218514364)(54,4.869617269565357)(55,4.200032517332225)(56,5.065347628059856)(57,4.267785017157263)(58,4.040180477817513)(59,4.916762520062454)(60,5.07207628747146)(61,5.100503504263237)(62,5.006247191269027)(63,5.025874846834533)(64,5.021982084865931)(65,5.020432882403752)(66,5.489893561511791)(67,5.254830125706172)(68,5.123265061354198)(69,5.330348072288001)(70,5.525516352097222)(71,5.565428598207376)(72,5.5744556680107245)(73,5.749804397495545)(74,5.695299567735811)(75,5.74607173224959)(76,5.734482238489979)(77,5.6805781510354745)(78,5.9465808999196605)(79,5.61911266655256)(80,5.719856521709892)(81,5.630988194285678)(82,5.738237317860829)(83,5.745826048257088)(84,5.799241623207911)(85,5.771740146224141)(86,5.718017711460385)(87,5.7098245077994285)(88,5.517324024458433)(89,5.7794307684847315)(90,5.753582894735635)(91,5.486295512933738)(92,5.888277668905905)(93,5.559802541379517)(94,5.589827047796176)(95,5.739148207265323)(96,5.6346800801037125)(97,5.301096310277565)(98,5.716921336114012)(99,5.523397143591319)(100,5.561309727935055)
};
            
            \addplot[
                color=blue,
                mark=square,
                mark size=1pt
                ]
                coordinates {(1,1.6923625115081382)(2,3.2468790697836587)(3,3.039356526350466)(4,3.237386670983203)(5,3.249111224100173)(6,3.231761323395017)(7,3.941045245337232)(8,4.4370315752137675)(9,3.3743737164700773)(10,3.4286138765086154)(11,3.3220714443078503)(12,3.4832077514385196)(13,3.210104215228669)(14,3.1309442574255306)(15,3.1999571700400122)(16,2.8835869905388396)(17,3.1162102522057435)(18,2.9960761194693077)(19,3.2212022354550296)(20,3.30353730598661)(21,3.3115849009104497)(22,3.4388987544541045)(23,3.662790266570643)(24,3.746965868271718)(25,3.3811405425237253)(26,3.6181178307222908)(27,3.774852510765493)(28,3.946762171363799)(29,4.069554796862665)(30,3.7871940868480185)(31,3.708624731038059)(32,4.2412377459224615)(33,4.002865373541182)(34,4.296189759966131)(35,4.559264852164265)(36,4.069986110436264)(37,4.276621958571617)(38,4.252215749470987)(39,4.859794054911514)(40,4.629198698796959)(41,4.772691601387489)(42,4.891567665213179)(43,5.070878629173551)(44,4.810808787786355)(45,5.197381473082965)(46,5.3099391328317616)(47,5.573305614949147)(48,5.624704627909551)(49,5.054415563009531)(50,5.30214098897652)(51,5.5395482788397095)(52,5.221043774756156)(53,5.549333332715747)(54,5.053363393231212)(55,5.235446533409553)(56,5.573012706951719)(57,5.411987989941968)(58,5.702787349236823)(59,5.45420703810151)(60,5.875905191508409)(61,5.712056703000107)(62,5.9380872303617505)(63,5.915630248065624)(64,6.151056846606238)(65,5.988879517655825)(66,6.005701123713175)(67,6.157217995824419)(68,6.170016347198127)(69,6.158222616444999)(70,5.934890625423916)(71,6.180273638538351)(72,6.311231210429933)(73,6.492970872158122)(74,6.218597651373481)(75,6.3172485187192144)(76,6.525074761007753)(77,6.492004081507392)(78,6.534281567236453)(79,6.423243482317402)(80,6.472910338801359)(81,6.639550987148636)(82,6.467041313568645)(83,6.671275145728612)(84,6.698534643459066)(85,6.679340205450002)(86,6.660819125479868)(87,6.739374173489291)(88,6.761725296543661)(89,6.778813584853156)(90,6.777393858447095)(91,6.729542499082349)(92,6.6466552224134565)(93,6.804811010224237)(94,6.806984221784072)(95,6.790208737745145)(96,6.770353432510979)(97,6.791751545882512)(98,6.781505783375735)(99,6.815036745669845)(100,6.791357569257231)
};

            \addplot[
                color=yellow,
                mark=diamond,
                mark size=1pt
                ]
                coordinates {(1,2.5459664101713653)(2,3.224542452432763)(3,3.00745726702051)(4,3.1630271929844653)(5,3.4037695462219855)(6,3.7022310060079633)(7,4.036625499937819)(8,3.599236046838188)(9,3.558435323520082)(10,3.9037940657743615)(11,4.358568393658892)(12,4.0389519803156375)(13,3.567154347180365)(14,3.9066584087778953)(15,3.6506035832839916)(16,3.4190010951262453)(17,3.716720371025904)(18,3.5177697527193734)(19,3.839278786359864)(20,3.745314344763756)(21,4.121339844805853)(22,3.605568894865277)(23,3.799117803653187)(24,4.293627194294306)(25,4.631184909646915)(26,4.48823367747589)(27,4.562615581204957)(28,4.472183352000563)(29,4.261624099196834)(30,4.215503393250887)(31,4.403368918625549)(32,4.631537994690667)(33,5.171646382406015)(34,5.083155553036919)(35,5.1848657638709925)(36,5.839956030926016)(37,5.689674512994225)(38,5.395235924341172)(39,5.797230403815633)(40,5.883678064048688)(41,5.9674916545622985)(42,6.322174415748333)(43,6.0009263152472965)(44,5.78097597398212)(45,5.930149398674475)(46,6.372773282180163)(47,6.542988570444335)(48,6.469698913544615)(49,6.355255123292493)(50,6.342764013112387)(51,6.517037891348865)(52,6.644862148259207)(53,6.666724272072313)(54,6.59565788660651)(55,6.72975077002604)(56,6.682926357448977)(57,6.584948150751907)(58,6.432938324732678)(59,6.571630286139863)(60,6.717610837421686)(61,6.706835927946546)(62,6.730848153707424)(63,6.686628831741966)(64,6.747875601928605)(65,6.7834325752784315)(66,7.1204485003556695)(67,6.992856648003308)(68,6.897685945491606)(69,6.922012650172844)(70,7.102623151304405)(71,7.081248046280704)(72,7.029109379279121)(73,7.079720427877752)(74,7.013100041705712)(75,7.0992524129287595)(76,7.040262355903121)(77,7.046157243036144)(78,7.042535288052978)(79,7.068360594836111)(80,7.118035319988655)(81,7.170563076561857)(82,7.137644122296246)(83,7.133216988340081)(84,7.130164152401711)(85,7.119689194856084)(86,7.126740437602808)(87,7.2039545328419745)(88,7.1950899834865245)(89,7.13598539088374)(90,7.123012811204141)(91,7.1496336155326095)(92,7.099544210451468)(93,7.21301054795371)(94,7.192042444612538)(95,7.170301979788156)(96,7.144487288371426)(97,7.1912262536217275)(98,7.150127060243857)(99,7.233615701344208)(100,7.166869307710268)
};
            \legend{micro, mini, small}
            \end{axis}
        \end{tikzpicture}
        \caption{METR-LA}
    \end{subfigure}
    \caption{Validation and training loss differences of the best-performing models on the respective training and validation dataset. Larger models tend to overfit on both datasets.}
    \label{fig:train_val_diff_plot}
\end{figure}
While the model appears to overfit on METR-LA by its increasing validation loss, the validation metrics are actually monotonously decreasing. This could be caused by the presence of missing values in the dataset. In Figure ~\ref{fig:metr-la_viz}, the traffic speed across all sensors abruptly drops to zero at approximately hours 8, 30, and 34. Such abrupt drops are absent in the PEMS-BAY dataset. Although we did not directly download the dataset, the simultaneous changes across all sensors in the METR-LA dataset strongly indicate missing values rather than traffic jams or road blockages. To alleviate the effect of missing values, we impute the training dataset with historical averages. Overall, these missing values increase the task difficulty.
\begin{figure}[tb]
    \centering
    \includegraphics[width=\linewidth]{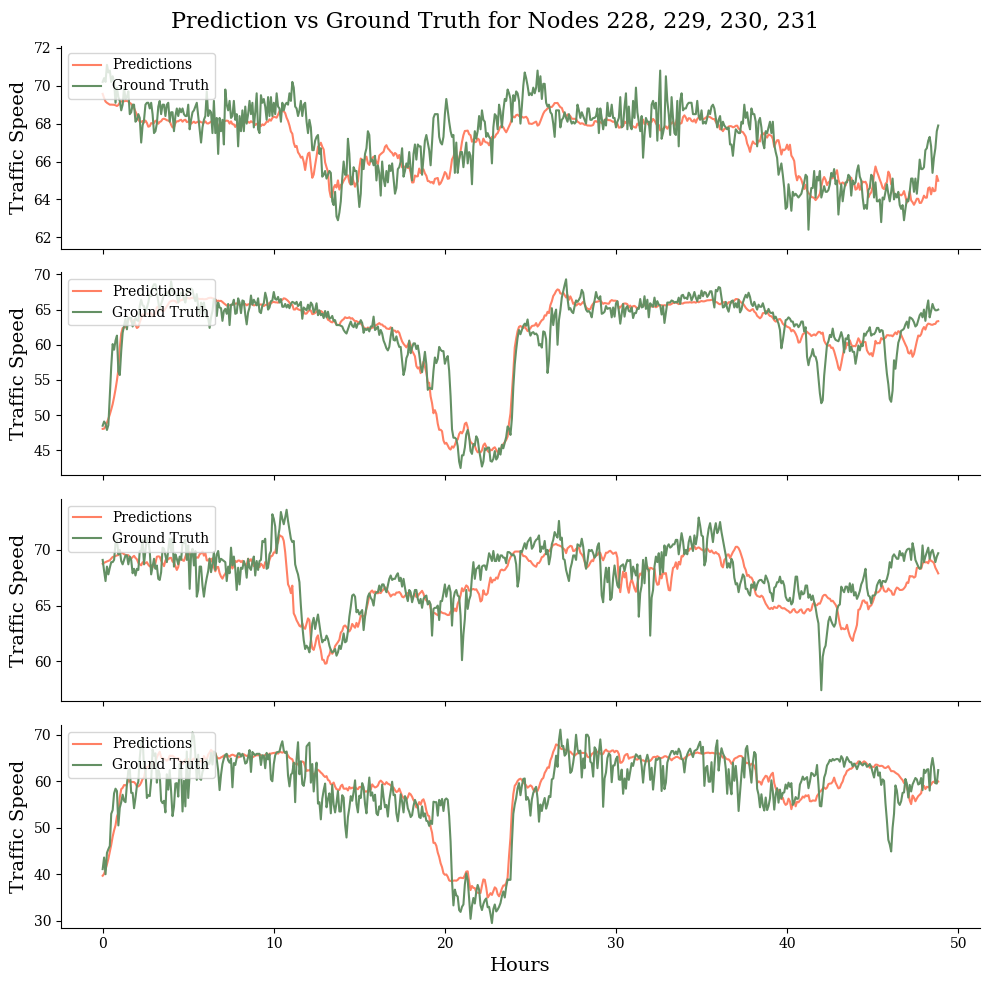}
    \caption{Prediction visualizations of T-Graphormer on PEMS-BAY. In the selected sensors shown above, the traffic speed drops (congestion) at different times of the day. Their congestion behaviours are also different. T-Graphormer can learn these complex traffic patterns.}
    \label{fig:pems-bay_viz}
\end{figure}

\begin{figure}[tb]
    \centering
    \includegraphics[width=\linewidth]{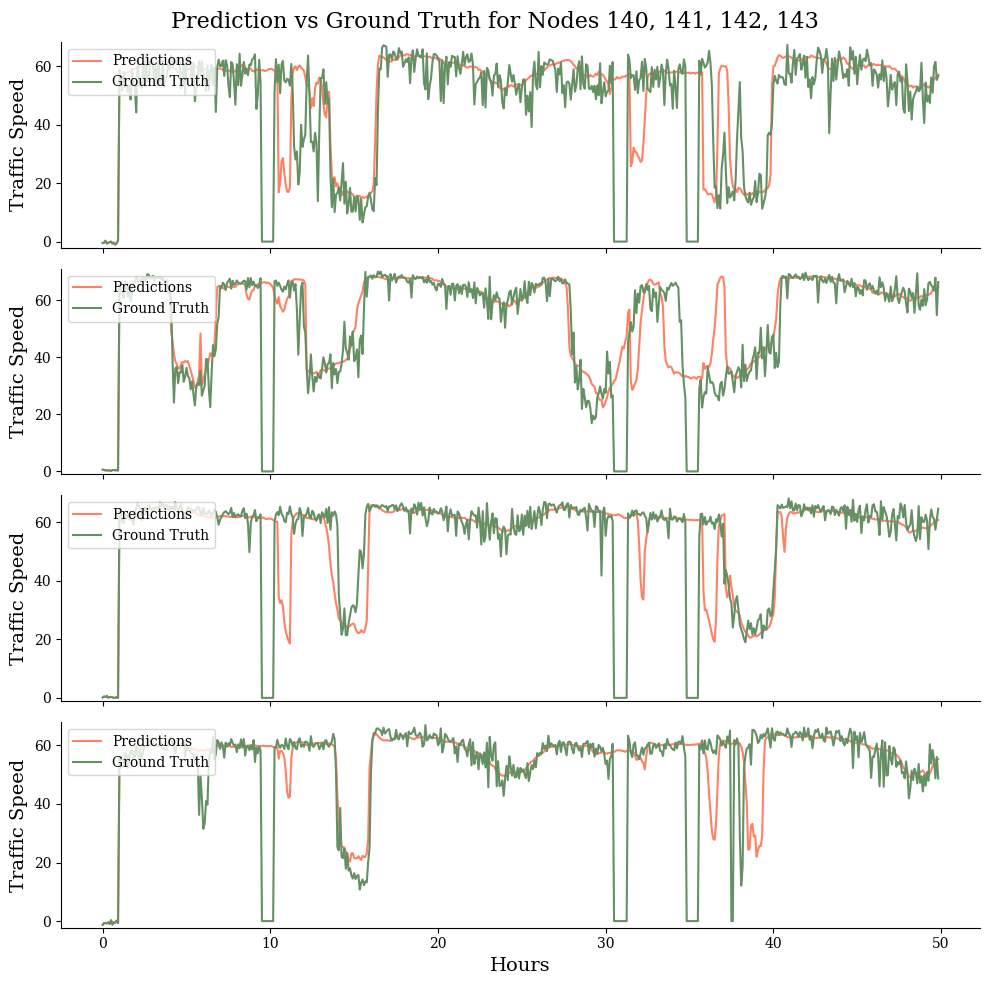}
    \caption{Prediction visualizations of T-Graphormer on METR-LA.}
    \label{fig:metr-la_viz}
\end{figure}
\subsection{Additional Settings}\label{sec: add_implement}
\begin{table}[hbt]
    \centering
        \begin{tabular}{@{}llllll@{}}
        \toprule
        Datasets & \# Sensors & \# Samples & Sampling Rate & Mean & Std \\ \midrule
        PEMS-BAY & 325        & 52116      & 5 minute &  54.40 & 19.49    \\
        METR-LA  & 207        & 34727      & 5 minute & 62.73 & 9.44     \\ 
        \bottomrule
        \end{tabular}
    \caption{Details of the dataset used for evaluation. The mean and standard deviation (std) are calculated across time and space.}
    \label{tab:dataset}
\end{table}
To construct the sensor graph, we compute the pairwise road network distances between sensors and build the adjacency matrix using thresholded Gaussian kernel \citep{li2018diffusion}. $W_{i,j} = \exp \left(
- \frac{\mathrm{dist}(v_i, v_j)^2}{\sigma^2}
\right)$ if $\mathrm{dist}(v_i, v_j) \leq \kappa$, otherwise 0, where $W_{i,j}$ is the edge distance between nodes $i, j$, and $\mathrm{dist}$ measures the physical distance between the nodes. $\sigma$ is the standard deviation of the distances, and $\kappa$ is the threshold for determining if an edge exists. 

We also use the training dataset to $Z$-score normalize the entire dataset. In cases where the time-of-day information is available for the dataset, we concatenate it to the traffic speed measurement to enrich the input.

Besides the details mentioned in the main text, we also use gradient accumulation to increase batch size. This results in an effective batch size that is calculated by multiplying the original batch size by the number of GPUs (graphics processing units) used in distributed data parallelism and the number of accumulated gradient iterations. We also utilize layer-wise learning rate decay following \citep{raffel2020exploring}, where the learning rate decays exponentially in earlier layers. This is typically used to stabilize training when fine-tuning Transformer models. We find that using gradient clipping and dropout improves T-Graphormer training as well \citep{srivastava2014dropout}.

All training has been done on SLURM workload manager environments. Training was done on 4 compute nodes. Each compute node has 187 gigabytes of memory, 2 Intel Silver 4216 Cascade CPUs, and 4 NVIDIA V100 Volta GPUs with 32 gigabytes of memory. On the PEMS-BAY dataset, training, validation, and model checkpoint took an average of 4.37 hours for micro-sized models, 7.5 hours for mini-sized models, and 13.8 hours for small-sized models. On the METR-LA dataset, it took an average of 2.77 hours for micro-sized models, 4.5 hours for mini-sized models, and 8.5 hours for small-sized models.  

\subsection{Additional Datasets}
To demonstrate the generalizability of our method, we evaluated it on three additional traffic prediction datasets introduced by \citet{song2020spatial} (see Table~\ref{tab:pems_dataset}).
\begin{table}[hbt]
    \centering
    \begin{tabular}{@{}llllll@{}}
    \toprule
    Datasets & \# Sensors & \# Samples & Sampling Rate & Mean & Std \\ \midrule
    PEMS03 & 358 & 26185 & 5 minute & 181.38 & 144.41 \\
    PEMS04 & 307 & 16969 & 5 minute & 207.23 & 156.48 \\
    PEMS08 & 170 & 17833 & 5 minute & 229.86 & 145.62 \\
    \bottomrule
    \end{tabular}
    \caption{Details of the additional dataset used for evaluation.}
    \label{tab:pems_dataset}
\end{table}
The data preprocessing pipeline is similar to that used for PEMS-BAY and METR-LA, with the main difference being the data split ratio, which is set to $6:2:2$ for these datasets. We report the prediction results for a horizon of 12 in Table~\ref{tab:pems_results}. For all three datasets, we trained the smallest variant of T-Graphormer. The optimal hyperparameters for each dataset are detailed in Table~\ref{tab:pems_config}.
\begin{table*}[hbt]
\centering
\resizebox{0.95\linewidth}{!}{%
\begin{tabular}{lllllllllll}
\toprule
\multicolumn{1}{l}{\multirow{2}{*}{\textbf{Method}}} & \multicolumn{3}{c}{\textbf{PEMS03}} & \multicolumn{3}{c}{\textbf{PEMS04}} & \multicolumn{3}{c}{\textbf{PEMS08}} \\ \cmidrule(lr){2-4}\cmidrule(lr){5-7}\cmidrule(lr){8-10} 
& MAE & RMSE & MAPE & MAE & RMSE & MAPE & MAE & RMSE & MAPE \\ \midrule
{VAR} & 23.65 & 38.26 & {24.51} & 23.75 & 36.66 & {18.09} & 23.46 & 36.33 & 15.42 \\
{FC-LSTM} & 21.33 & 35.11 & {23.33} & 27.14 & 41.59 & {18.20} & 22.20 & 34.06 & 14.20 \\
{DCRNN} & 18.18 & 30.31 & {18.91} & 24.70 & 38.12 & {17.12} & 17.86 & 27.83 & 11.45 \\
{STGCN} & 17.49 & 30.12 & {17.15} & 22.70 & 35.55 & {14.59} & 18.02 & 27.82 & 11.40 \\
{Graph WaveNet} & 19.85 & 32.94 & {19.31} & 25.45 & 39.70 & {17.29} & 19.13 & 31.05 & 12.68 \\
{ASTGCN} & 17.69 & 29.66 & {19.40} & 22.93 & 35.22 & {16.56} & 18.61 & 28.16 & 13.08 \\
{STSGCN} & 17.48 & 29.21 & {16.78} & 21.19 & 33.65 & {13.90} & 17.13 & 26.80 & \textbf{10.96} \\\midrule

{\ac{T-Graphormer}} & \textbf{15.88} & \textbf{24.58} & {\textbf{15.62}} & \textbf{20.40} & \textbf{30.37} & {\textbf{13.86}} & \textbf{16.88} & \textbf{24.61} & 11.43 \\ \bottomrule
\end{tabular}
}
\caption{Horizon 12 forecasting results for the PEMS03, PEMS04, and PEMS08 datasets. We report the test metrics from the best configuration of T-Graphormer. The best-performing model for each metric is bolded.}
\label{tab:pems_results}
\end{table*}
\begin{table}[htb]
    \centering
    \begin{tabular}{cccc}\toprule
\multirow{2}{*}{\textbf{Configuration}} & \multicolumn{3}{c}{\textbf{Dataset}}                                       \\ \cmidrule(lr){2-4} & PEMS03    & PEMS04   & PEMS08  \\\midrule
optimizer                               & \multicolumn{3}{c}{AdamW} \\
optimizer momentum                      & \multicolumn{3}{c}{$\beta_1,\beta_2=0.9, 0.999$}                         \\
learning rate schedule                  & \multicolumn{3}{c}{cosine decay}   \\
epochs                                  & 100     & 100     & 100             \\
learning rate                           & 3.50e-3 & 7.50e-3 & 4.50e-3     \\
gradient clipping                       & 1.0     & 1.0     & 1.0            \\
weigh decay                             & 1e-5    & 1e-4    & 1e-6         \\
warmup epochs                           & 10      & 10     & 10               \\
batch size                              & 128     & 128     & 128             \\
dropout                                 & 0.1     & 0.1     & 0.1            \\
layer-wise decay                        & 0.90    & 0.90    & 0.90    \\
\# of parameters (M) & 1.77 & 1.69 & 1.48  \\\bottomrule
\end{tabular}
    \caption{Best training hyperparameters of T-Graphormer mini on the additional datasets.}
    \label{tab:pems_config}
\end{table}

On the PEMS03 dataset, T-Graphormer achieved improvements of \textbf{9.15\%} in MAE, \textbf{15.85\%} in RMSE, and \textbf{6.91\%} in MAPE. On PEMS04, it improved MAE by \textbf{3.73\%} and RMSE by \textbf{9.75\%}. For PEMS08, T-Graphormer enhanced RMSE by \textbf{8.17\%}. These results indicate a consistent trend: as the dataset size increases, T-Graphormer demonstrates stronger predictive performance. Notably, since the model is trained with the Huber loss with a high $\delta$, the most significant improvement is observed in the RMSE metric.

The baselines used for these datasets differ from those for PEMS-BAY and METR-LA due to variations in preprocessing and incomplete results reported in the original manuscripts of STEP, PDFormer, and STAEformer. Specifically, STEP reports results only for PEMS04, while PDFormer and STAEformer omit results for PEMS03, limiting the ability to validate their performance. We do not report results for the PEMS07 dataset because we are unable to fit the entire flattened graph sequence ($883\times 12 = 10596$ tokens) into the memory of our GPUs during training. 
\end{document}